\documentclass[preprint,12pt]{elsarticle}
\usepackage{hyperref}
\usepackage{breakurl} 
\usepackage{tabularx}
\usepackage{subcaption}
\usepackage{longtable}
\usepackage{multirow}
\usepackage{makecell}
\usepackage{amsmath}
\usepackage{float}
\usepackage{comment}
\usepackage{natbib}
\usepackage{algorithm}
\usepackage{algorithmic}

\begin{document}

\begin{frontmatter}

\title{MedPrompt: LLM-CNN Fusion with Weight Routing for Medical Image Segmentation and Classification} %

\author{Shadman Sobhan\textsuperscript{1}}\corref{cor1}
\ead{shadmansobhan114@gmail.com}
\author{Kazi Abrar Mahmud\textsuperscript{2}}
\ead{abrar.mahmud790011@gmail.com}
\author{Abduz Zami\textsuperscript{3}}
\ead{abduz.zami@gmail.com}
\cortext[cor1]{Corresponding author.}

\address{\textsuperscript{1*,2} Department of Electrical and Electronic Engineering,\\Bangladesh University of Engineering and Technology, Dhaka-1000, Bangladesh}
\address{\textsuperscript{3}Department of Computer Science \& Engineering,\\Rajshahi University of Engineering \& Technology, Rajshahi-6204, Bangladesh}

\begin{abstract}

Current medical image analysis systems are typically task-specific, requiring separate models for classification and segmentation, and lack the flexibility to support user-defined workflows. To address these challenges, we introduce MedPrompt, a unified framework that combines a few-shot prompted Large Language Model (Llama-4-17B) for high-level task planning with a modular Convolutional Neural Network (DeepFusionLab) for low-level image processing. The LLM interprets user instructions and generates structured output to dynamically route task-specific pretrained weights. This \textit{weight routing} approach avoids retraining the entire framework when adding new tasks—only task-specific weights are required, enhancing scalability and deployment. We evaluated MedPrompt across 19 public datasets, covering 12 tasks spanning 5 imaging modalities. The system achieves a 97\% end-to-end correctness in interpreting and executing prompt-driven instructions, with an average inference latency of ~2.5 seconds, making it suitable for near real-time applications. DeepFusionLab achieves competitive segmentation accuracy (e.g., Dice ~0.9856 on lungs) and strong classification performance (F1 ~0.9744 on tuberculosis). Overall, MedPrompt enables scalable, prompt-driven medical imaging by combining the interpretability of LLMs with the efficiency of modular CNNs.

\end{abstract}


\end{frontmatter}

\section{Introduction}
\label{Intro}
Natural language is the most widely used medium for human communication. Conversely, medical diagnostics predominantly rely on imaging modalities like X-rays, CT, MRI, and endoscopy \citep{weissleder2015advancing}, \citep{ballard2021role}. Convolutional Neural Networks (CNNs), designed for image-based tasks, have become standard in medical image analysis \citep{salehi2023study}, \cite{suzuki2017overview}, \cite{sarvamangala2022convolutional}. Medical image analysis fundamentally relies on two key tasks: segmentation, which localizes specific regions such as organs or lesions, and classification, which identifies diseases \citep{chowdhary2020segmentation}, \citep{yang2021artificial}, \citep{ma2020diagnostic}. 

Despite the rise of multi-task vision models, few architectures enable segmentation from user-instructed natural language prompts. Most prompt-based segmentation models, such as SAM (Segment Anything Model) \citep{kirillov2023segment} and its fine-tuned variant Med-SAM \citep{ma2024segment}, are generalized, meaning they can segment a wide variety of anatomical structures without needing task-specific retraining. But such generalized models have three major problems. First, they rely on spatial prompts (e.g., points or bounding boxes), which are less intuitive and harder to automate than natural language. Second, fine-tuning them for new anatomy requires large-scale data and computing, making them inflexible. And the final one is that these prompt-based models are generally segmentation-focused. Fine-tuning is required for these models to learn classification. 

At present, Large Language Models can understand texts to a great extent and even continue a conversation like a real human being \citep{li2025fundamental}, \citep{matarazzo2025survey}. Also, Vision Language Models (VLMs) are the state-of-the-art models that are proficient in understanding both images and text \citep{ghosh2024exploring}. Although recent vision-language models demonstrate strong multimodal understanding, they lack direct integration with image processing backbones, limiting their use in tasks like segmentation.

In this work, we propose MedPrompt, an LLM-CNN unified framework where the LLM parses the user prompt and queries a weight database containing task-specific CNN weights trained with DeepFusionLab. DeepFusionLab switches to a segmentation or classification model by dynamically selecting and loading pretrained weights (‘weight routing’) based on the LLM’s structured response. Weight Routing allows for easy scalability, as adding a new task only requires training DeepFusionLab to generate corresponding weights and registering them. 

To summarize, our core contributions are as follows:
\begin{itemize}
    \item We propose a unified LLM-CNN framework, MedPrompt, capable of executing both segmentation and classification from natural language prompts using a single model, DeepFusionLab.
    \item The system supports complex, multi-stage, and conditional queries.
    \item We introduce dynamic weight routing to enable scalable task expansion without full-model retraining.
    \item MedPrompt achieves real-time performance with an average latency of ~2.5 seconds, making it practical for clinical use.
\end{itemize}

The rest of the paper is structured as follows: Section \ref{Lit} presents a comprehensive review of related works and analyzes existing research gaps. Section \ref{Met} discusses the framework in detail, DeepFusionLab architecture. In section \ref{Exp}, we explore the datasets used in the study, the experimental setup, which includes different hyperparameter tunings, and training procedures in our study. Section \ref{Eval} presents our experimental results and compares the results between our model and other works. The section also discusses the ablation study and the performance of the whole framework. The paper concludes with a summary of our findings and potential future directions in Section \ref{con}.

\section{Literature Review}
\label{Lit}

Convolutional Neural Networks (CNNs) are widely applied in medical image classification. Architectures like ResNet \citep{he2016deep}, DenseNet \citep{huang2017densely}, EfficientNet \citep{tan2019efficientnet}, and Vision Transformers (ViT) \citep{dosovitskiy2020image} have advanced feature learning and efficiency. These models have been adapted to detect diseases such as COVID-19 \citep{monshi2021covidxraynet}, eye conditions \citep{muthukannan2022optimized}, and pneumonia using CheXNet \citep{rajpurkar2017chexnet}.

Similar to classification, CNNs also revolutionized segmentation. Early encoder-decoder models such as FCN \citep{long2015fully} and SegNet \citep{badrinarayanan2017segnet} enabled pixel-wise prediction. U-Net \citep{ronneberger2015u}, with its symmetric skip connections, became widely adopted, while DeepLabv3+ \citep{chen2018encoder} improved multi-scale feature capture. More recently, hybrid models like TransUNet \citep{chen2021transunet} and Swin-Unet \citep{cao2022swin} combine transformers with CNNs, achieving better global context and efficiency across tasks like tumor \citep{bhandari2020convolutional} and vessel segmentation \citep{l2017recent}.

Even though CNNs were traditionally proposed for either segmentation or classification, some models jointly perform classification and segmentation to improve efficiency. Mask R-CNN \citep{he2017mask} enables simultaneous detection and segmentation. Multi-task networks like Swin UNETR \citep{grzeszczyk2022multi} and COVID-CT-Mask-Net \citep{ter2022covid} combine transformer or CNN backbones with dual heads for integrated diagnosis.

Promptable segmentation models like SAM \citep{kirillov2023segment} use spatial cues such as points or boxes, while MedSAM \citep{ma2024segment} adapts SAM for medical imaging through domain-specific fine-tuning. More advanced models like SEEM \citep{zou2023segment}, LISA \citep{lai2024lisa}, and GLaMM \citep{rasheed2024glamm} incorporate multimodal prompts and attention mechanisms to support open-world or few-shot segmentation.

In contrast, vision-language models (VLMs) enable zero-shot classification and retrieval by aligning images with text. MedCLIP \citep{wang2022medclip} and BioViL \citep{boecking2022making} enhance multimodal understanding via contrastive learning and domain-specific pretraining. Hybrid models such as MedCLIP-SAM \citep{koleilat2024medclip} and LLaVA-Med \citep{li2023llava} bridge vision and language for text-prompted segmentation. Meanwhile, large-scale models like BiomedGPT \citep{luo2023biomedgpt} and GPT-4V are being explored for multitask clinical applications.

Recent work in vision-language models has focused on multitask, instruction-driven systems. VisionLLM v2\citep{wu2024visionllm}, for example, uses a “super link” to connect an LLM with task-specific decoders, enabling unified handling of tasks like VQA, localization, and image editing. SAM4MLLM \citep{chen2024sam4mllm} combines Segment Anything (SAM) with an LLM for referring segmentation, while Zeus \citep{dai2025zeus} uses a frozen LLM to produce segmentation instructions from medical images. InstructSeg \citep{wei2024instructseg} unify referring and reasoning-based segmentation through text-guided prompts. These systems rely on fixed vision backbones (often SAM or CLIP) and use LLMs mainly for high-level prompting.

The reviewed literature highlights substantial progress in medical image analysis, yet several limitations hinder the clinical deployment of current models.

Most CNN-based models are designed for either classification (e.g., ResNet, CheXNet) or segmentation (e.g., U-Net, DeepLabv3+), requiring multiple models for different tasks and complicating clinical workflows. Prompt-based models like SAM and MedSAM support segmentation via spatial cues but do not handle classification. Conversely, vision-language models (VLMs) such as MedCLIP and LLaVA-Med excel at classification but lack segmentation capability, limiting their use in tasks requiring anatomical localization.

Few systems support natural language prompts, which are more intuitive for clinicians. Additionally, adapting existing models to new organs or diseases typically demands costly fine-tuning. Most models are rigid, lacking support for conditional or multi-step workflows, and often suffer from high inference latency, limiting real-time clinical use.

MedPrompt addresses these gaps through a unified, instruction-driven design. A single CNN backbone, DeepFusionLab, handles both classification and segmentation, guided by natural language instructions interpreted by a Large Language Model (LLM). MedPrompt introduces a dynamic weight routing mechanism that loads task-specific pretrained weights based on the LLM’s structured plan. This enables easy extension to new tasks without architectural changes. The system supports context-aware, multi-step pipelines and loads only the necessary weights, improving efficiency and scalability for domain-specific applications.

\section{Methodology}
\label{Met}

\begin{figure}[h]
    \centering
    \includegraphics[width=1\textwidth]{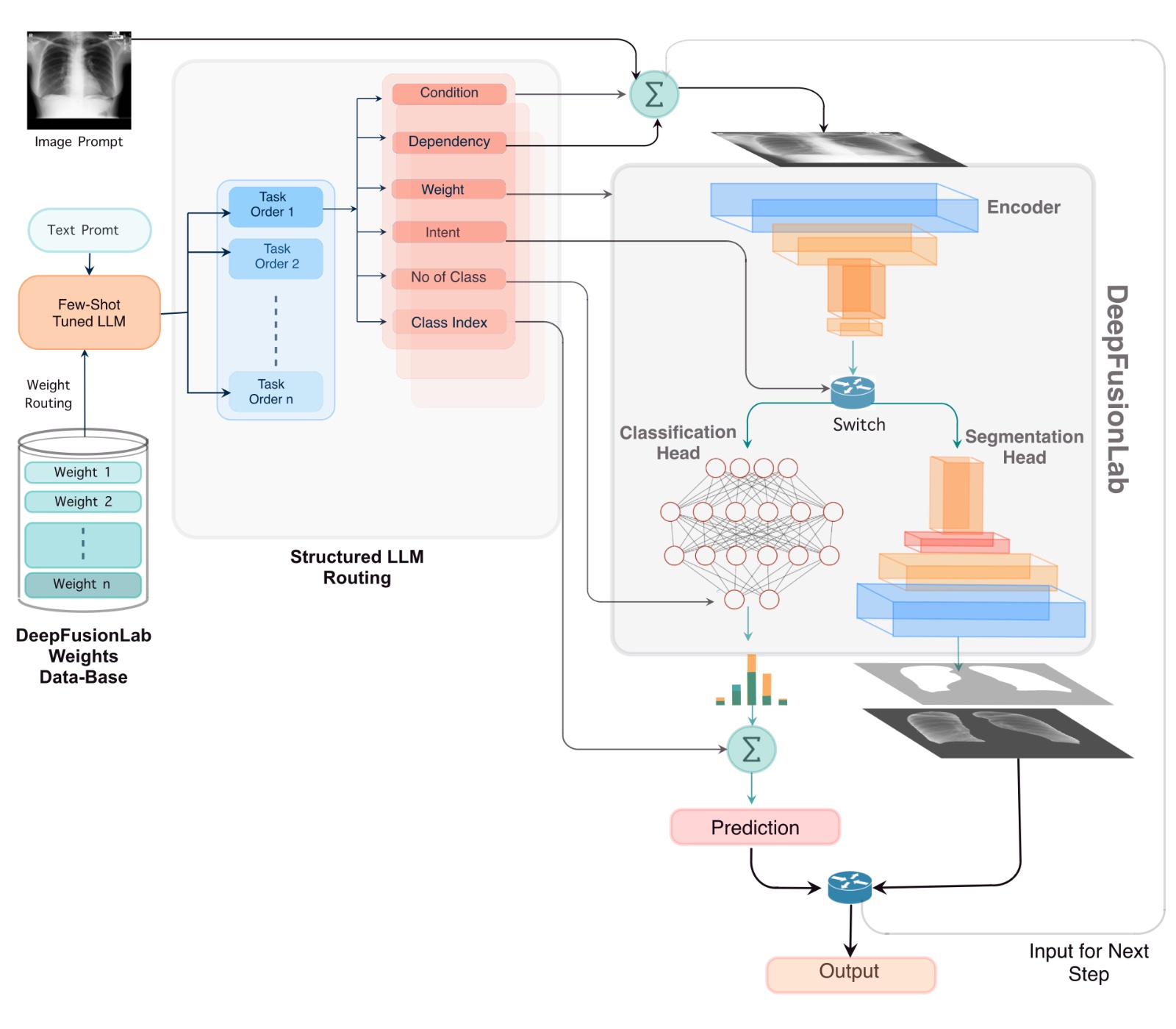}
    \caption{MedPrompt Architecture}
    \label{MedPrompt}
\end{figure}

The overall architecture of the proposed framework is given in Figure \ref{MedPrompt}

The methodology is organized into three sub-sections for clarity and ease of explanation. These components are: (I) Inputs of the framework, (II) Language model-based text processing and structured output generation, and (III) Coordination of structured input with DeepFusionLab.  

\subsection{Inputs of the Framework:}

The proposed framework, MedPrompt, always requires a text input, an image, and access to the Weight Database.

The text input or prompt is a natural language-based instruction that specifies the type of operation, of which organ/disease the user wants to perform. Some simple prompts can be: 

- "Isolate the lung region."

- "Do you find pneumonia in the given image?"

- "This is a person's report. Does the person have malaria?"
\\[2\baselineskip]  
The weight database consists of model weights produced through training using the DeepFusionLab framework.  For accurate interpretation by large language models (LLMs), these weights should follow a standardized naming convention. The ideal naming format is shown in Table \ref{tab:Weight_Naming}

\begin{table}[h]
\caption{Standardized Weight Naming Format}
\label{tab:Weight_Naming}
\begin{tabular}{|l|l|l|}
\hline
Type of Weight                                                                    & Standard Format                                                                    & Example                           \\ \hline
\multirow{2}{*}{Segmentation}                                                     & \multirow{2}{*}{Seg\_Organ/Disease\_Modality}                                      & Seg\_Lung\_Chest X-ray            \\ \cline{3-3} 
                                                                                  &                                                                                    & Segmentation\_Lung\_CXR           \\ \hline
\multirow{2}{*}{\begin{tabular}[c]{@{}l@{}}Binary \\ Classification\end{tabular}} & \multirow{2}{*}{Cls\_Disease\_Modality}                                            & Cls\_TB\_Chest X-ray              \\ \cline{3-3} 
                                                                                  &                                                                                    & Classification\_Tuberculosis\_CXR \\ \hline
\begin{tabular}[c]{@{}l@{}}Multi Class \\ Classification\end{tabular}             & \begin{tabular}[c]{@{}l@{}}Cls\_Disease 1- Disease 2-...\\ \_Modality\end{tabular} & Cls\_Covid-Pneumonia\_CXR         \\ \hline
\end{tabular}
\end{table}

It should be noted that strict adherence to this naming convention is not mandatory. Variants such as "Cls", "Class", or "Classification" are all acceptable, provided the overall format "\textbf{\texttt{intent}\texttt{\_target}\texttt{\_modality}}" remains clear and semantically consistent.

The third input is the medical image, on which the segmentation or classification will be performed. 

\subsection{Language Model-Based Text Processing and Structured Output Generation}

Figure \ref{JSON} demonstrates the methodology for structured output generation by LLM.
\begin{figure}[h]
    \centering
    \includegraphics[width=1\textwidth]{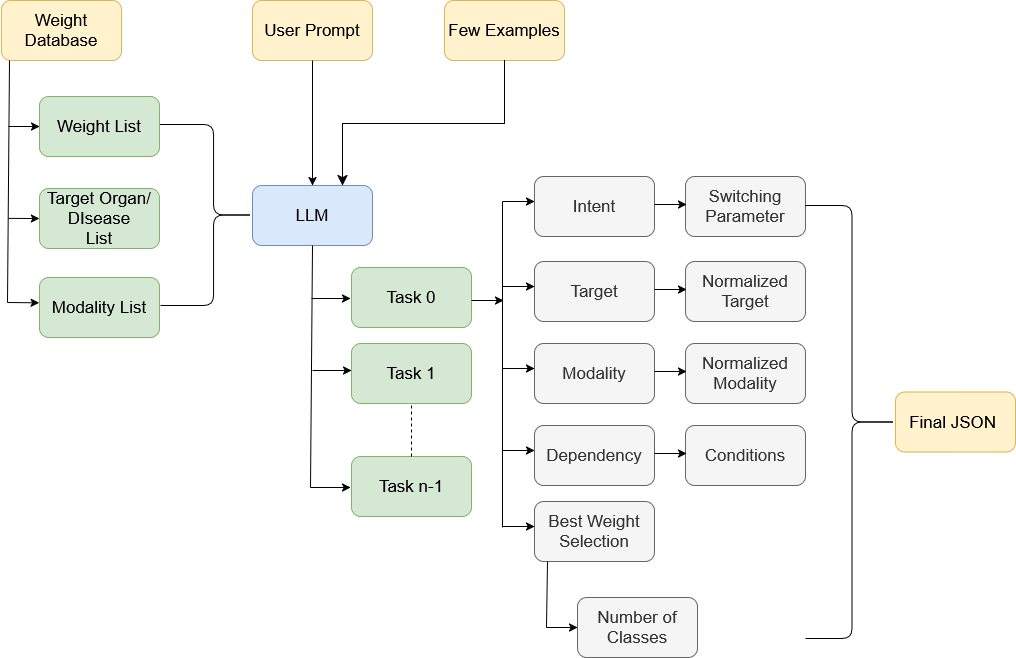}
    \caption{Text Processing and Structured Output Generation}
    \label{JSON}
\end{figure}

This begins by retrieving all available pretrained model weights from a centralized weight database. Each weight filename encodes three attributes: the task intent (\textit{classification} or \textit{segmentation}), the target organ or condition, and the imaging modality. These metadata elements are extracted and used to compile reference lists of available targets and modalities.

A few-shot prompted Large Language Model (LLM) interprets the user's natural language prompt and decomposes it into a sequence of structured tasks. Each task is defined by an \textit{intent}, \textit{target}, and optionally, a \textit{modality}. These fields are then normalized against the reference lists derived from the weight database. We used \textbf{Llama-4-17B} as the LLM here.

Target and modality normalization is performed in two stages. First, a rule-based dictionary resolves common variations and synonyms (e.g., ``tuberculosis'' $\rightarrow$ ``tb''). For cases not covered by the dictionary, the system falls back to LLM-based semantic normalization. For example, a query containing the term ``phthisis'' is semantically matched to tuberculosis. This hybrid approach ensures robustness and flexibility in parsing diverse clinical expressions. A similar normalization strategy is applied to the modality field when it is specified.

The framework also supports conditional execution. If the user prompt implies that a task should be performed only under certain conditions—such as following a positive or negative outcome of a prior task—the system encodes these dependencies. 

Once the tasks are parsed and normalized, the framework performs \textit{weight routing}, which assigns the most appropriate pretrained model to each task. This is achieved by computing a \textit{match score} for each candidate weight based on three criteria: intent match, target similarity, and modality similarity. Semantic similarities are calculated using cosine similarity between Sentence-BERT embeddings of the task and weight descriptors.

The final score \( S \) for a given weight is computed as:

\begin{equation}
S = I + \alpha \cdot \text{sim}_{\text{target}} + \beta \cdot \text{sim}_{\text{modality}}
\end{equation}

where:
\begin{itemize}
    \item \( I = 1 \) if the task intent exactly matches the weight’s intent, else \( I = 0 \),
    \item \( \text{sim}_{\text{target}} \in [0, 1] \) is the cosine similarity between normalized task and weight targets,
    \item \( \text{sim}_{\text{modality}} \in [0, 1] \) is the cosine similarity between normalized task and weight modalities,
    \item \( \alpha = 1.5 \) and \( \beta = 1.0 \) 
\end{itemize}

We set the target similarity weight \( \alpha = 1.5 \) to give more importance to target matching than modality when selecting pretrained weights. This value was chosen by testing different values in the range \( [1.2, 2.0] \) on a validation set of user prompts. The value \( \alpha = 1.5 \) gave the most accurate and consistent matches.

The maximum score is:

\begin{equation}
S_{\text{max}} = 1 + 1.5 \cdot 1 + 1.0 \cdot 1 = 3.5
\end{equation}

A weight is considered a valid match only if its score exceeds a predefined threshold of 1.6. The threshold for accepting a weight was set to \( S > 1.6 \). We evaluated score distributions and weight selection outcomes across annotated prompt-task pairs. A threshold of 1.6 reliably filtered out weak or spurious matches, while retaining all correct ones. Since the maximum score is 3.5, this threshold allows partial matches (e.g., correct target but uncertain modality) while avoiding unrelated weights. Lower thresholds were found to increase false matches, especially when only modality was similar.

When multiple classification weights exist for a given target, the system typically prefers binary models over multiclass ones. This preference emerges naturally due to higher semantic similarity scores; binary classifiers tend to focus on a single condition and often exhibit superior diagnostic performance.

For classification tasks, the number of classes and their corresponding labels are inferred from the filename or an associated JSON mapping file. For segmentation tasks, a default two-class setup is assumed, representing the background and the target region.

The final pipeline is returned as a structured JSON object, with each task containing its resolved intent, target, modality, dependencies, conditional logic (if any), selected weight, and class-level information.

\subsection{DeepFusionLab Architecture:}
In the proposed methodology, we employed DeepFusionLab, which is a convolutional neural network architecture capable of handling both classification and segmentation tasks using a shared encoder and task-specific decoding branches. 

\begin{figure}[h]
    \centering
    \includegraphics[width=1\textwidth]{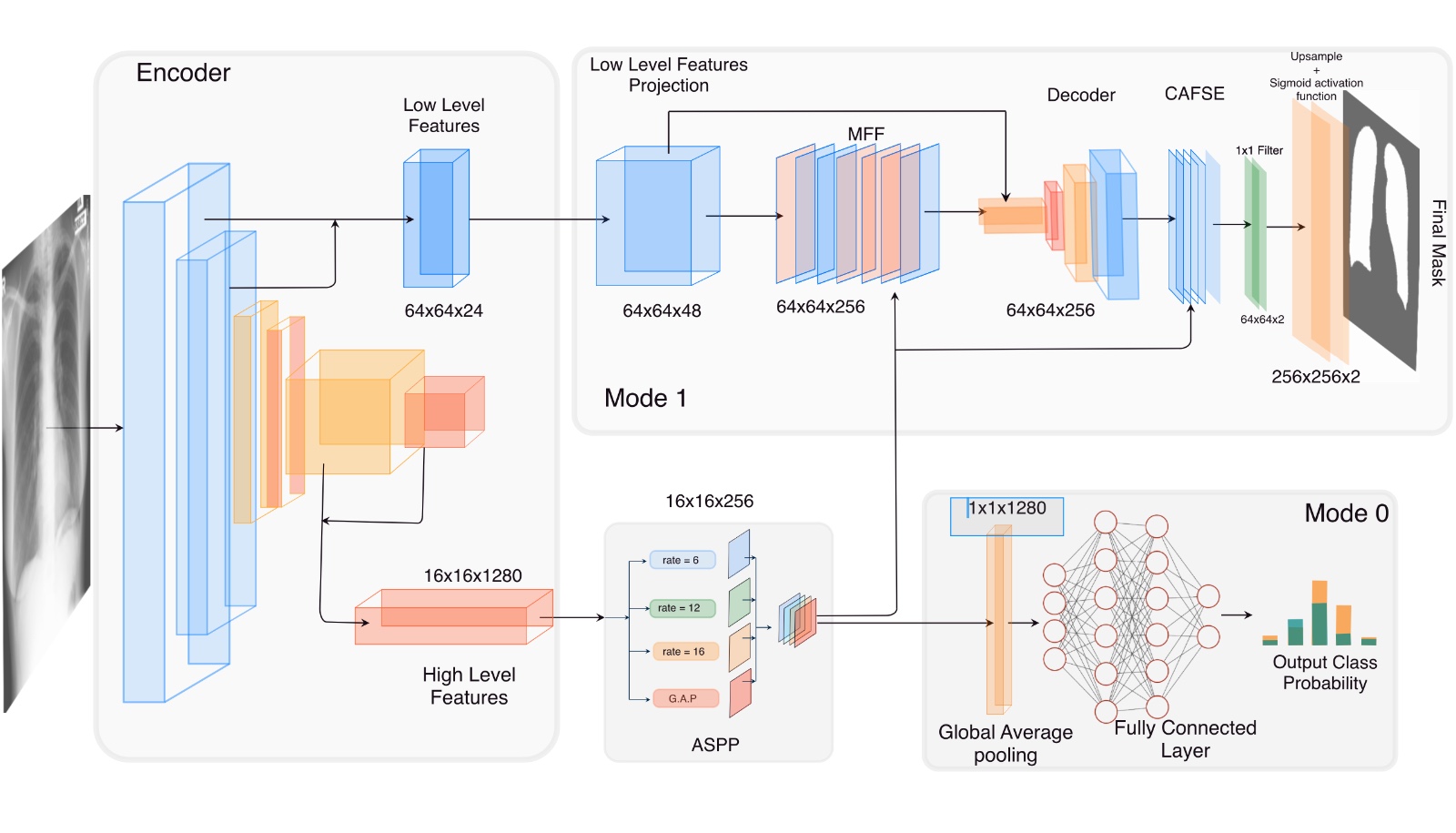}
    \caption{Architecture of DeepFusionLab}
    \label{DFL}
\end{figure}

Figure \ref{DFL} shows the architecture of DeepFusionLab.

The model takes an input image of shape 256×256×3 and adapts its behavior based on a binary mode flag: Mode = 0 for classification and Mode = 1 for segmentation.

\subsubsection{Encoder (Feature Extractor)}
The encoder uses EfficientNet-B0 pretrained on ImageNet as the backbone. EfficientNet-B0 was chosen because of its computational efficiency and accuracy. Two levels of features are extracted:
\begin{itemize}
    \item \textbf{Low-Level Features:} Early layers in the network capture fine details and produce a feature map of size 64×64 with 24 channels. These details are useful for segmentation.
    \item \textbf{High-Level Features:} Deeper layers capture more abstract, semantic information and output a feature map of size 16×16 with 1280 channels. These features are useful for both classification and segmentation.
\end{itemize}

\subsubsection{Classification Pathway}
When the mode is set to 0, the model performs classification. The high-level features go through an Atrous Spatial Pyramid Pooling (ASPP) module. Global Pooling is applied, and the feature map is flattened to be converted into a single vector that summarizes the whole image. This vector is fed through a dense layer with ReLU activation and dropout for regularization, followed by a final dense layer that outputs scores for each class. Using softmax activation, the scores are turned into probabilities indicating the likelihood of each class.

\subsubsection{Segmentation Pathway}
From the EfficientNet-B0 encoder, low and high-level features are extracted. The low-level features are passed through 48 1×1 convolutions to increase the number of channels from 24 to 48, making them more expressive. The low-level features are thus converted into low-level features projections. In parallel, the high-level features are passed through the same ASPP module. These two outputs are then combined using a Multiscale Feature Fusion (MFF) block. The fused feature map, along with low-level feature projections, is then passed through a decoder. To further improve the result, a Coarse and Fine Step Enhancement (CAFSE) block combines the decoder output and ASPP features using channel attention. Finally, 1×1 convolution reduces the output to two channels representing the binary segmentation mask, which is then upsampled to the original input size using bilinear interpolation. A sigmoid activation function is applied to produce the final probability mask, indicating the likelihood of each pixel belonging to the target region.

\subsubsection{ASPP Block Architecture}
\begin{figure}[h]
    \centering
    \includegraphics[width=1\textwidth]{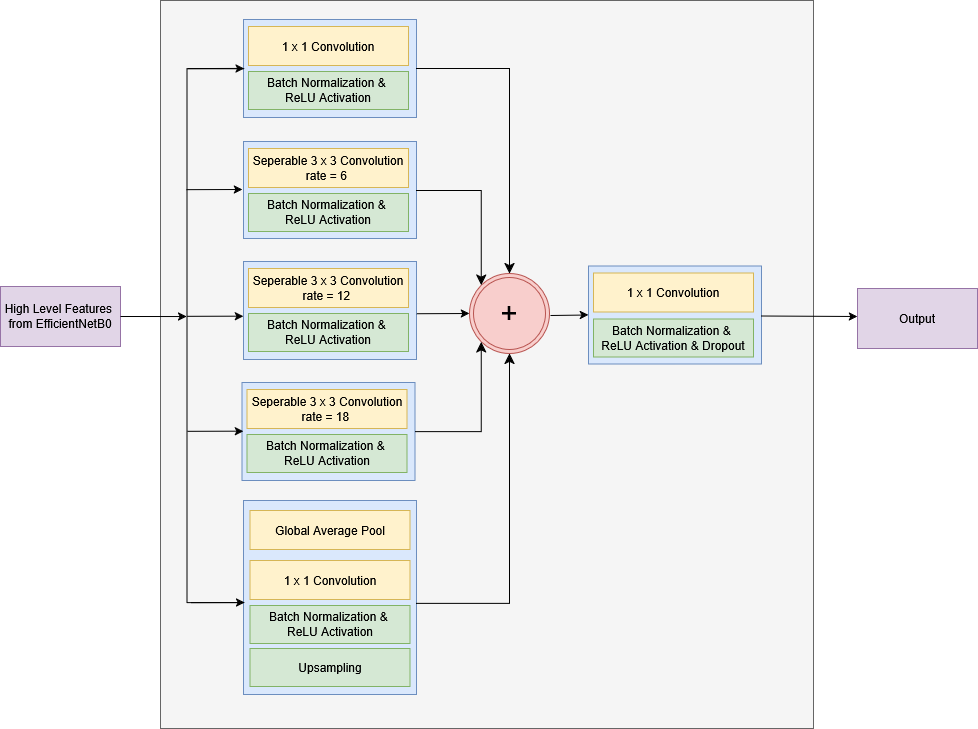}
    \caption{Atrous Spatial Pyramid Pooling Block Architecture}
    \label{ASPP}
\end{figure}

The Atrous Spatial Pyramid Pooling (ASPP) block is designed to extract rich multi-scale contextual information by applying multiple convolution operations with different dilation rates. As shown in Figure~\ref{ASPP}, the block consists of parallel branches: a $1 \times 1$ convolution, multiple $3 \times 3$ depthwise separable convolutions with different atrous (dilation) rates, and a global average pooling branch followed by a $1 \times 1$ convolution. 
This is followed by a $1 \times 1$ projection convolution to unify the channel dimension and a dropout layer to prevent overfitting. \textbf{The ASPP block captures both local and global context efficiently}.

\subsubsection{MFF Block Architecture}
\begin{figure}[h]
    \centering
    \includegraphics[width=1\textwidth]{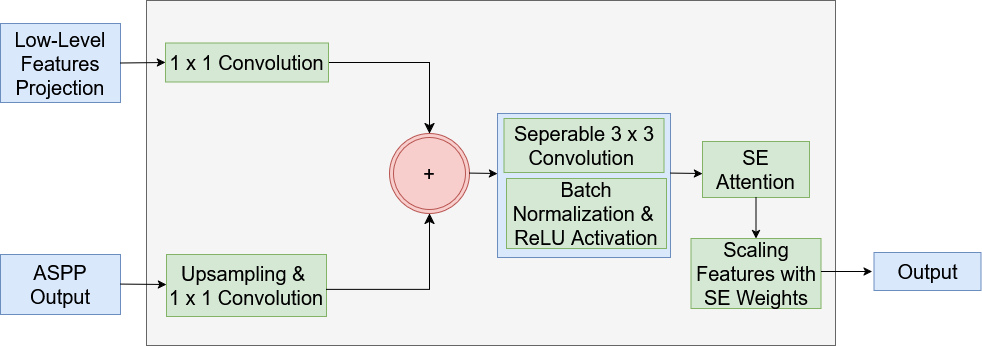}
    \caption{Multi Feature Fusion Block Architecture}
    \label{MFF}
\end{figure}

The Multi Feature Fusion (MFF) block, illustrated in Figure~\ref{MFF}, is designed to combine high-level and low-level features extracted from different stages of the encoder.
The high-level features, usually from the ASPP output, are first upsampled to match the spatial resolution of the low-level features. Both low-level and high-level features are projected using $1 \times 1$ convolutions, summed, then passed through a $3 \times 3$ convolution, batch normalization, and ReLU. Finally, an SE block applies channel attention. To further enhance, a channel-wise attention mechanism is implemented via a Squeeze-and-Excitation (SE) block. MFF adaptively emphasizes important channels and suppresses less useful ones through the SE mechanism.

\subsubsection{Decoder Architecture}
\begin{figure}[h]
    \centering
    \includegraphics[width=1\textwidth]{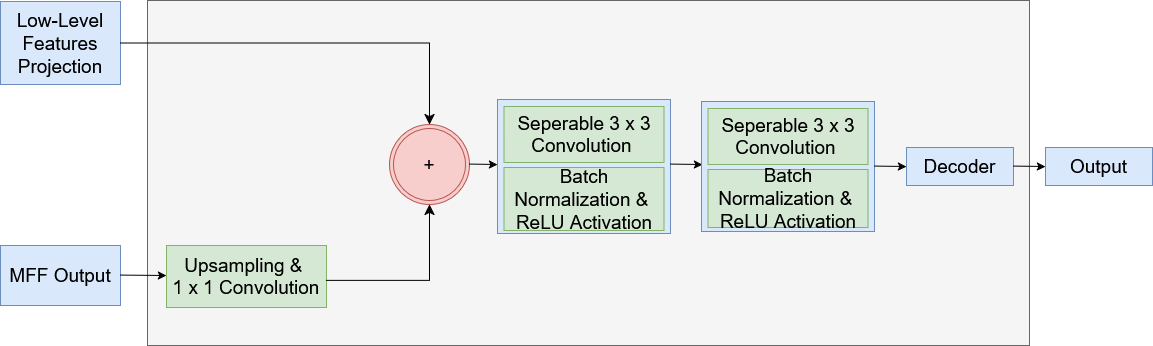}
    \caption{Decoder Architecture}
    \label{Decoder}
\end{figure}

The decoder, shown in Figure~\ref{Decoder}, progressively recovers spatial details lost during encoding.
It first upsamples the input feature map (from the MFF block) to match the spatial resolution of the low-level features. A $1 \times 1$ convolution is applied to the upsampled features, followed by batch normalization and ReLU. These processed features are then concatenated with the low-level features extracted from early layers of the encoder. The combined features are refined through two depthwise separable convolutions interleaved with batch normalization, ReLU activation, and dropout. 

\subsubsection{CAFSE Architecture}
\begin{figure}[h]
    \centering
    \includegraphics[width=1\textwidth]{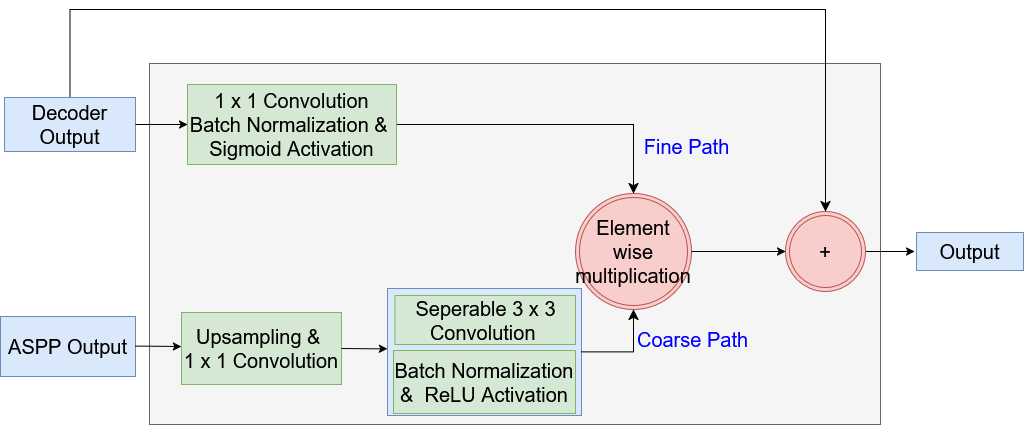}
    \caption{Coarse and Fine Step Enhancement Block Architecture}
    \label{CAFSE}
\end{figure}

The Coarse and Fine Step Enhancement (CAFSE) block, as illustrated in Figure~\ref{CAFSE}, fuses semantic features from the ASPP block (coarse) with spatially detailed features from the decoder (fine).
The coarse stage, output from ASPP, is rich in semantics but low in resolution. The fine stage, output from the decoder, is rich in spatial detail but lacks broader context.

The ASPP output is first upsampled to the resolution of the decoder output. It then passes through a $1 \times 1$ convolution and a separable $3 \times 3$ convolution to enhance coarse semantic features. Simultaneously, the decoder output undergoes a $1 \times 1$ convolution followed by a sigmoid activation to generate an attention map. This attention map modulates the enhanced coarse features through element-wise multiplication, which is then added to the original decoder features. The result is a contextually enhanced and spatially refined feature map. CAFSE effectively balances semantic abstraction with spatial precision, which is especially valuable in medical image segmentation.

In DeepFusionLab, the separable convolution was applied instead of the general convolution to reduce computational complexity, making the model lightweight yet effective.

Table \ref{tab:Blocks} summarizes the parameter count and operational purpose of each component in the proposed DeepFusionLab

\begin{table}[]
\caption{Functional roles and parameter count of each major block in the proposed DeepFusionLab architecture.}
\label{tab:Blocks}
\begin{tabular}{|l|l|l|}
\hline
\textbf{Block}                                                     & \textbf{Number of Parameters} & \textbf{Roles}                                                                                                \\ \hline
\begin{tabular}[c]{@{}l@{}}Backbone\\ EfficientNet-B0\end{tabular} & 4.007 M                       & \begin{tabular}[c]{@{}l@{}}Extracts low and high-level \\ features from the input image.\end{tabular}         \\ \hline
ASPP                                                               & 2.003 M                       & \begin{tabular}[c]{@{}l@{}}Captures multi-scale context using \\ parallel atrous convolutions.\end{tabular}   \\ \hline
MFF                                                                & 0.684 M                       & \begin{tabular}[c]{@{}l@{}}Fuses low-level and high-level features \\ for spatial refinement.\end{tabular}    \\ \hline
Decoder                                                            & 0.106 M                       & \begin{tabular}[c]{@{}l@{}}Reconstruct spatial details \\ for mask generation.\end{tabular}    \\ \hline
CAFSE                                                              & 0.656 M                       & \begin{tabular}[c]{@{}l@{}}Enhances spatial precision through \\ coarse-to-fine feature integration.\end{tabular} \\ \hline
Classification Head                                                & 0.656 M                       & \begin{tabular}[c]{@{}l@{}}Maps global features to \\ final class probabilities.\end{tabular}                 \\ \hline
Segmentation Head                                                  & 257                           & \begin{tabular}[c]{@{}l@{}}Produces pixel-level prediction \\ maps for the target structure.\end{tabular}     \\ \hline
\textbf{Overall}                                                   & \textbf{8.12 M}               & ---                                                                                                           \\ \hline
\end{tabular}
\end{table}

\section{Experimental Setup}
\label{Exp}

The success of a deep learning model depends heavily on the careful design and execution of the experimental setup. In this section, we outline the various steps taken to prepare, train, and evaluate our model. 

\subsection{Datasets Used:}
The datasets used in this study to train DeepFusionLab are shown in Table \ref{tab:dataset}
\begin{table}[h!]
\caption{Datasets Trained}
\label{tab:dataset}
\begin{tabular}{|l|l|l|l|}
\hline
Purpose                                                                                                    & Dataset                                                                  & Modality                           & \begin{tabular}[c]{@{}l@{}}Training/Testing\\ Images\end{tabular} \\ \hline
Lung Segmentation                                                                                         & JSRT \citep{JSRT}                                                                    & \multirow{5}{*}{Chest X-ray}       & 197/50                                                            \\ \cline{1-2} \cline{4-4} 
\begin{tabular}[c]{@{}l@{}}Lung Segmentation\\ \& Tuberculosis\\ Classification\end{tabular}              & Montgomery \citep{Montgomery}                                                             &                                    & 110/28                                                            \\ \cline{1-2} \cline{4-4} 
\multirow{2}{*}{\begin{tabular}[c]{@{}l@{}}Tuberculosis\\ Classification\end{tabular}}                    & Shenzhen \citep{Shenzhen}                                                                &                                    & 529/133                                                           \\ \cline{2-2} \cline{4-4} 
                                                                                                          & CXR\_dataset \citep{rahman2020reliable}                                                            &                                    & 3360/840                                                          \\ \cline{1-2} \cline{4-4} 
\begin{tabular}[c]{@{}l@{}}Covid \& Viral \\ Pneumonia \\ Classification\end{tabular}                     & \begin{tabular}[c]{@{}l@{}}Covid-19\\ Radiography\\ Dataset \citep{chowdhury2020can}\end{tabular} &                                    & 16930/4235                                                        \\ \hline
\multirow{3}{*}{\begin{tabular}[c]{@{}l@{}}Retinal Blood \\ Vessel \\ Segmentation\end{tabular}}          & Drive  \citep{staal2004ridge}                                                                  & \multirow{3}{*}{Color fundus}      & 20/20                                                             \\ \cline{2-2} \cline{4-4} 
                                                                                                          & Stare \citep{hoover2000locating}                                                                    &                                    & 16/4                                                              \\ \cline{2-2} \cline{4-4} 
                                                                                                          & Chase\_DB1 \citep{fraz2012ensemble}                                                               &                                    & 22/6                                                              \\ \hline
\multirow{2}{*}{\begin{tabular}[c]{@{}l@{}}Colon-Polyp\\ Segmentation\end{tabular}}                       & KVASIR\_SEG \citep{jha2019kvasir}                                                            & \multirow{2}{*}{Endoscopy}         & 800/200                                                           \\ \cline{2-2} \cline{4-4} 
                                                                                                          & CVC\_ClinicDB \citep{bernal2015wm}                                                          &                                    & 489/123                                                           \\ \hline
\multirow{5}{*}{\begin{tabular}[c]{@{}l@{}}Optic-cup \&\\ Optic-disc segmentation\end{tabular}}           & CRFO \citep{kiefer2023catalog}                                                                    & \multirow{7}{*}{Color fundus}      & 35/9                                                              \\ \cline{2-2} \cline{4-4} 
                                                                                                          & Drishti \citep{sivaswamy2015comprehensive}                                                                 &                                    & 80/21                                                             \\ \cline{2-2} \cline{4-4} 
                                                                                                          & G1020 \citep{velpulafundus}                                                                   &                                    & 632/159                                                           \\ \cline{2-2} \cline{4-4} 
                                                                                                          & Origa \citep{zhang2010origa}                                                                    &                                    & 520/130                                                           \\ \cline{2-2} \cline{4-4} 
                                                                                                          & Papila \citep{kovalyk2022papila}                                                                  &                                    & 390/98                                                            \\ \cline{1-2} \cline{4-4} 
\begin{tabular}[c]{@{}l@{}}Optic-cup, Optic-disc\\ segmentation \& Glaucoma\\ classification\end{tabular} & Refuge1 \citep{orlando2020refuge}                                                                 &                                    & 640/160                                                           \\ \cline{1-2} \cline{4-4} 
Glaucoma Classification                                                                                   & Rim-One \citep{fumero2011rim}                                                                 &                                    & 425/107                                                           \\ \hline
Malaria Classification                                                                                    & \begin{tabular}[c]{@{}l@{}}NIH Malaria\\ Dataset \citep{rajaraman2018pre}\end{tabular}            & \multirow{2}{*}{Microscopy images} & 11023/2756                                                        \\ \cline{1-2} \cline{4-4} 
Leukemia Classification                                                                                   & \begin{tabular}[c]{@{}l@{}}C-NMC\_\\ Leukemia \citep{gupta2022c}\end{tabular}               &                                    & 8528/2133                                                         \\ \hline
\end{tabular}
\end{table}

\subsection{DeepFusionLab Training}

The dataset was initially split into training and testing sets in an 80:20 ratio in such a way that every class of data is divided into that ratio. The training set was later split into training and validation sets in a 90:10 ratio. 

Setting the model to segmentation mode, the model was trained for segmentation for 200 epochs. The batch size was adjusted based on the size of the training dataset: a batch size of 8 was used for datasets with fewer than 500 training images, while a batch size of 16 was used otherwise. In some cases, a batch size of 32 was selected depending on dataset characteristics and available GPU memory. For optimization during training, Dice Loss was used as the loss function. Dice Loss was derived from the Dice Coefficient. To prevent overfitting, real-time data augmentation was applied during training that included transformations such as horizontal and vertical flips.

The model was first set to classification mode to train on classification datasets and then trained for 100 epochs with a batch size of 32 or 64, or 128 if the number of training set images was less than 1000; otherwise, it was set to 64. For much larger datasets, the batch size was fixed at 128. As the loss function, we implemented weighted loss, and similar to segmentation training, real-time data augmentation, such as horizontal and vertical flips, was applied during training to prevent overfitting. 

To keep the pipeline simple and focus on evaluating the model’s performance under minimal preprocessing, we applied only image resizing and histogram equalization. All images were resized to \(256 \times 256\) to ensure uniform input dimensions and compatibility with the network architecture. 

\subsection{Hyperparameter Tuning}

To achieve optimal performance and efficient training, we carefully tuned several hyperparameters. 

For segmentation, the Adam optimizer was used with an initial learning rate of 0.001. A \textit{CosineAnnealingWarmRestarts} scheduler was employed to adjust the learning rate cyclically. It was configured with \( T_0 = 1 \), meaning the first learning rate cycle lasted for one epoch, and \( T_{\text{mult}} = 2 \), indicating that each subsequent cycle was twice as long as the previous one (i.e., 1, 2, 4, 8, ... epochs). The minimum learning rate was set to \( \eta_{\min} = 1 \times 10^{-9} \). Early stopping was applied based on validation loss, with a patience of 30 epochs and a minimum improvement threshold of \( 1 \times 10^{-10} \). Model checkpointing was used to save the model whenever an improvement in the validation Dice coefficient was observed.

For classification, the model was compiled using the Adam optimizer with a learning rate of 0.001. During training, early stopping was applied based on validation loss with a patience of 30 epochs to prevent overfitting. Additionally, model checkpointing was used to save the weights whenever an improvement in validation accuracy was observed, ensuring the best-performing model was retained for final evaluation.

\subsection{LLM Selection}
One of the critical steps in this work is the selection of a Large Language Model (LLM) to generate structured outputs. We selected Llama-4-Maverick-17B due to its being an open-source LLM, strong language understanding, and efficient instruction-following capabilities. No fine-tuning was performed; instead, we used few-shot prompting, providing 10 examples within the prompt to guide the model's behavior. Importantly, none of these examples were included in the evaluation set. For prompts, at first task description was provided along with how to reason explicitly for each JSON component. After that, 10 example questions and their answer were provided. LLM inference was performed in this study via Groq APIs.

\subsection{Machine Configuration}

This study was conducted using the computational resources provided by Kaggle’s GPU-accelerated platform. A Kaggle Notebook environment equipped with two NVIDIA Tesla T4 GPUs, each offering 16GB of GDDR6 memory and 2,560 CUDA cores, was utilized. 

\section{Evaluation \& Discussion}
\label{Eval}
In this section, we discussed the performance of the proposed pipeline. In this study, our evaluation can be separated into 3 segments. The first one is the accuracy of structured output generation, DeepFusionLab performance, and the overall framework assessment. As the structured output and framework assessment are closely related, they are shown in a single Table.

\subsection{DeepFusionLab Performance Across Datasets:}
The proposed CNN, DeepFusionLab, is capable of both classification and segmentation. So, the performance evaluation of DeepFusionLab involves evaluating both classification and segmentation datasets.

\subsubsection{Evaluation on Classification Datasets:}
Table \ref{tab:classification-results} demonstrates the performance of the proposed CNN on seven targets with 8 different datasets (10, including combined datasets).
\begin{longtable}{|l|l|l|l|l|l|l|l|l|l|}
\caption{Evaluation of DeepFusionLab across Classification Datasets} \label{tab:classification-results} \\
\hline
Target & Dataset & Accuracy & Precision & Recall & F1 & AUC & \begin{tabular}[c]{@{}l@{}}Training\\ Time\\(minute)\end{tabular} \\
\hline
\endfirsthead

\hline
Target & Dataset & Accuracy & Precision & Recall & f1 & AUC & \begin{tabular}[c]{@{}l@{}}Training\\ Time\end{tabular} \\
\hline
\endhead

\multirow{4}{*}{Tuberculosis}                                                       & CXR\_Dataset                                                                & \begin{tabular}[c]{@{}l@{}}0.9995 \\ ±\\ 0.0007\end{tabular}  & \begin{tabular}[c]{@{}l@{}}0.9924 \\ ±\\  0.0005\end{tabular}   & \begin{tabular}[c]{@{}l@{}}0.9922 \\ ±\\  0.0224\end{tabular} & \begin{tabular}[c]{@{}l@{}}0.9923 \\ ±\\  0.0223\end{tabular} & \begin{tabular}[c]{@{}l@{}}1.0000 \\ ±\\  0.0000\end{tabular} & \begin{tabular}[c]{@{}l@{}}7.83 \\ ±\\ 1.5297\end{tabular}        \\ \cline{2-8} 
                                                                                    & Shenzhen                                                                    & \begin{tabular}[c]{@{}l@{}}0.8972 \\ ± \\ 0.0197\end{tabular} & \begin{tabular}[c]{@{}l@{}}0.8993\\  ±\\  0.0182\end{tabular}   & \begin{tabular}[c]{@{}l@{}}0.8975\\  ±\\  0.0196\end{tabular} & \begin{tabular}[c]{@{}l@{}}0.8971 \\ ±\\ 0.0198\end{tabular}  & \begin{tabular}[c]{@{}l@{}}0.9442 \\ ±\\  0.0122\end{tabular} & \begin{tabular}[c]{@{}l@{}}2.7033 \\ ±\\  0.8571\end{tabular}     \\ \cline{2-8} 
                                                                                    & \begin{tabular}[c]{@{}l@{}}Shenzhen+\\ Montgomery\end{tabular}              & \begin{tabular}[c]{@{}l@{}}0.8985\\  ±\\  0.0127\end{tabular} & \begin{tabular}[c]{@{}l@{}}0.9008\\  ±\\  0.0100\end{tabular}   & \begin{tabular}[c]{@{}l@{}}0.8991 \\ ±\\  0.0123\end{tabular} & \begin{tabular}[c]{@{}l@{}}0.8984 \\ ±\\  0.0128\end{tabular} & \begin{tabular}[c]{@{}l@{}}0.9434\\  ±\\  0.0029\end{tabular} & \begin{tabular}[c]{@{}l@{}}2.6167\\  ±\\  0.0939\end{tabular}     \\ \cline{2-8} 
                                                                                    & Combined                                                                    & \begin{tabular}[c]{@{}l@{}}0.9825 \\ ±\\  0.0015\end{tabular} & \begin{tabular}[c]{@{}l@{}}0.9804 \\ ±\\  0.0026\end{tabular}   & \begin{tabular}[c]{@{}l@{}}0.9687 \\ ±\\  0.0017\end{tabular} & \begin{tabular}[c]{@{}l@{}}0.9744\\  ±\\  0.0021\end{tabular} & \begin{tabular}[c]{@{}l@{}}0.9986 \\ ±\\  0.0002\end{tabular} & \begin{tabular}[c]{@{}l@{}}16.9000 \\ ±\\  2.0500\end{tabular}    \\ \hline
Malaria                                                                             & \begin{tabular}[c]{@{}l@{}}NIH Malaria \\ Dataset\end{tabular}              & \begin{tabular}[c]{@{}l@{}}0.9776 \\ ±\\  0.0006\end{tabular} & \begin{tabular}[c]{@{}l@{}}0.9776 \\ ±\\  0.0006\end{tabular}   & \begin{tabular}[c]{@{}l@{}}0.9776\\ ±\\  0.0006\end{tabular}  & \begin{tabular}[c]{@{}l@{}}0.9776 \\ ±\\  0.0006\end{tabular} & \begin{tabular}[c]{@{}l@{}}0.9969\\  ±\\  0.0004\end{tabular} & \begin{tabular}[c]{@{}l@{}}72.7433 \\ ±\\  9.5022\end{tabular}  \\ \hline
Leukemia                                                                            & \begin{tabular}[c]{@{}l@{}}C-NMC\_\\ Leukemia\end{tabular}                  & \begin{tabular}[c]{@{}l@{}}0.9808 \\ ±\\  0.0013\end{tabular} & \begin{tabular}[c]{@{}l@{}}0.9802 \\ ±\\  0.0012\end{tabular}   & \begin{tabular}[c]{@{}l@{}}0.9754 \\ ±\\ 0.0023\end{tabular}  & \begin{tabular}[c]{@{}l@{}}0.9777 \\ ±\\ 0.0015\end{tabular}  & \begin{tabular}[c]{@{}l@{}}0.9959 \\ ±\\  0.0008\end{tabular} & \begin{tabular}[c]{@{}l@{}}63.8567\\  ±\\  7.4647\end{tabular}  \\ \hline
\multirow{3}{*}{Glaucoma}                                                           & REFUGE2                                                                     & \begin{tabular}[c]{@{}l@{}}0.9750\\  ±\\ 0.0102\end{tabular}  & \begin{tabular}[c]{@{}l@{}}0.9649 \\ ±\\ 0.0354\end{tabular}    & \begin{tabular}[c]{@{}l@{}}0.8935 \\ ±\\ 0.0312\end{tabular}  & \begin{tabular}[c]{@{}l@{}}0.9249 \\ ±\\ 0.0301\end{tabular}  & \begin{tabular}[c]{@{}l@{}}0.9630 \\ ±\\ 0.0166\end{tabular}  & \begin{tabular}[c]{@{}l@{}}4.5700 \\ ±\\ 0.1981\end{tabular}      \\ \cline{2-8} 
                                                                                    & RIM-ONE                                                                     & \begin{tabular}[c]{@{}l@{}}0.9031\\  ±\\ 0.0051\end{tabular}  & \begin{tabular}[c]{@{}l@{}}0.8911 \\ ±\\ 0.0050\end{tabular}    & \begin{tabular}[c]{@{}l@{}}0.9023 \\ ±\\ 0.0072\end{tabular}  & \begin{tabular}[c]{@{}l@{}}0.8960\\  ±\\  0.0058\end{tabular} & \begin{tabular}[c]{@{}l@{}}0.9628 \\ ±\\ 0.0009\end{tabular}  & \begin{tabular}[c]{@{}l@{}}10.5000 \\ ±\\ 5.4700\end{tabular}     \\ \cline{2-8} 
                                                                                    & Combined                                                                    & \begin{tabular}[c]{@{}l@{}}0.9363 \\ ±\\ 0.0070\end{tabular}  & \begin{tabular}[c]{@{}l@{}}0.9214\\  ±\\ 0.0132\end{tabular}    & \begin{tabular}[c]{@{}l@{}}0.9026 \\ ± \\ 0.0091\end{tabular} & \begin{tabular}[c]{@{}l@{}}0.9112 \\ ± \\ 0.0093\end{tabular} & \begin{tabular}[c]{@{}l@{}}0.9668\\ ± \\ 0.0013\end{tabular}  & \begin{tabular}[c]{@{}l@{}}19.8667 \\ ± \\ 4.3586\end{tabular}    \\ \hline
COVID                                                                               & \begin{tabular}[c]{@{}l@{}}COVID-19 \\ Radiography \\ Database\end{tabular} & \begin{tabular}[c]{@{}l@{}}0.9969 \\ ±\\  0.0005\end{tabular} & \begin{tabular}[c]{@{}l@{}}0.9967\\ \\  ±\\ 0.0005\end{tabular} & \begin{tabular}[c]{@{}l@{}}0.9952\\  ± \\ 0.0007\end{tabular} & \begin{tabular}[c]{@{}l@{}}0.9959 \\ ±\\ 0.0006\end{tabular}  & \begin{tabular}[c]{@{}l@{}}0.9996 \\ ±\\ 0.0004\end{tabular}  & \begin{tabular}[c]{@{}l@{}}90.6533 \\ ±\\  12.4130\end{tabular}  \\ \hline
\begin{tabular}[c]{@{}l@{}}Viral \\ Pneumonia\end{tabular}                          & \begin{tabular}[c]{@{}l@{}}COVID-19 \\ Radiography \\ Database\end{tabular} & \begin{tabular}[c]{@{}l@{}}0.9958 \\ ±\\ 0.0002\end{tabular}  & \begin{tabular}[c]{@{}l@{}}0.9906 \\ ±\\ 0.0008\end{tabular}    & \begin{tabular}[c]{@{}l@{}}0.9890 \\ ±\\ 0.0008\end{tabular}  & \begin{tabular}[c]{@{}l@{}}0.9898 \\ ±\\ 0.0005\end{tabular}  & \begin{tabular}[c]{@{}l@{}}0.9998 \\ ±\\  0.0001\end{tabular} & \begin{tabular}[c]{@{}l@{}}95.7233 \\ ±\\  16.15084\end{tabular} \\ \hline
\begin{tabular}[c]{@{}l@{}}COVID,\\ Viral \\ Pneumonia,\\ Lung Opacity\end{tabular} & \begin{tabular}[c]{@{}l@{}}COVID-19 \\ Radiography \\ Database\end{tabular} & \begin{tabular}[c]{@{}l@{}}0.9580 \\ ±\\ 0.0021\end{tabular}  & \begin{tabular}[c]{@{}l@{}}0.9671\\  ±\\ 0.0018\end{tabular}    & \begin{tabular}[c]{@{}l@{}}0.9619\\  ±\\ 0.0013\end{tabular}  & \begin{tabular}[c]{@{}l@{}}0.9645 \\ ±\\ 0.0013\end{tabular}  & N/A                                                           & \begin{tabular}[c]{@{}l@{}}112.3000 \\ ±\\  10.4539\end{tabular}  \\ \hline
\end{longtable}

From the table, it is evident that DeepFusionLab performed well across datasets. However, due to a smaller number of images in datasets like Shenzhen, Rim-One, the score was not as high as it was on other datasets. Training time varied mainly due to the execution of early stopping. When the image quality is good, like CXR\_Dataset, the training converges fast, and early stopping executes. But for not-so-good image quality, training does not converge that fast, and so it requires higher training time.

\subsubsection{Evaluation on Segmentation Datasets:}
Table \ref{tab:seg-metrics} demonstrates the performance of the proposed CNN on five targets with 19 different datasets (24, including combined datasets).

\begin{longtable}{|l|l|l|l|l|l|l|l|l|l|}
\caption{Evaluation of DeepFusionLab across Segmentation Datasets} \label{tab:seg-metrics}
\\ \hline
\makecell{Target} & \makecell{Dataset} & \makecell{Dice} & \makecell{IoU} & \makecell{AUC} & \makecell{Accura\\cy} & \makecell{Precisi\\on} & \makecell{Recall} & \makecell{F1} & \makecell{Training\\Time} \\
\hline
\endfirsthead

\hline
\makecell{Target} & \makecell{Dataset} & \makecell{Dice} & \makecell{IoU} & \makecell{AUC} & \makecell{Accura\\cy} & \makecell{Precisi\\on} & \makecell{Recall} & \makecell{F1} & \makecell{Training\\Time} \\
\hline
\endhead \hline
\multirow{3}{*}{Lung}                                                             & JSRT                                                       & \begin{tabular}[c]{@{}l@{}}0.9847 \\ ±\\ 0.0006\end{tabular} & \begin{tabular}[c]{@{}l@{}}0.9711 \\ ±\\ 0.001\end{tabular}  & \begin{tabular}[c]{@{}l@{}}0.9970 \\ ± \\ 0.0004\end{tabular} & \begin{tabular}[c]{@{}l@{}}0.9876 \\ ± \\ 0.0004\end{tabular} & \begin{tabular}[c]{@{}l@{}}0.9842\\ ±\\ 0.0001\end{tabular}   & \begin{tabular}[c]{@{}l@{}}0.9864 \\ ± \\ 0.0007\end{tabular} & \begin{tabular}[c]{@{}l@{}}0.9853 \\ ± \\ 0.0004\end{tabular} & \begin{tabular}[c]{@{}l@{}}8.0940 \\ ± \\ 1.0387\end{tabular} \\ \cline{2-10} 
                                                                                  & \begin{tabular}[c]{@{}l@{}}Montgo-\\ mery\end{tabular}     & \begin{tabular}[c]{@{}l@{}}0.9829 \\ ±\\ 0.0005\end{tabular} & \begin{tabular}[c]{@{}l@{}}0.9683 \\ ±\\ 0.0006\end{tabular} & \begin{tabular}[c]{@{}l@{}}0.9968 \\ ±\\ 0.0004\end{tabular}  & \begin{tabular}[c]{@{}l@{}}0.9883 \\ ±\\ 0.0009\end{tabular}  & \begin{tabular}[c]{@{}l@{}}0.9851 \\ ±\\ 0.0007\end{tabular} & \begin{tabular}[c]{@{}l@{}}0.9827 \\ ±\\ 0.0007\end{tabular}  & \begin{tabular}[c]{@{}l@{}}0.9837 \\ ±\\ 0.0007\end{tabular}  & \begin{tabular}[c]{@{}l@{}}6.2220 \\ ±\\ 0.5472\end{tabular}   \\ \cline{2-10} 
                                                                                  & Combined                                                   & \begin{tabular}[c]{@{}l@{}}0.9856 \\ ±\\ 0.0003\end{tabular} & \begin{tabular}[c]{@{}l@{}}0.9724 \\ ±\\ 0.0003\end{tabular} & \begin{tabular}[c]{@{}l@{}}0.9959 \\ ±\\ 0.0007\end{tabular}  & \begin{tabular}[c]{@{}l@{}}0.9887 \\ ±\\ 0.0004\end{tabular}  & \begin{tabular}[c]{@{}l@{}}0.9858 \\ ±\\ 0.0004\end{tabular} & \begin{tabular}[c]{@{}l@{}}0.9862 \\ ±\\ 0.0003\end{tabular}  & \begin{tabular}[c]{@{}l@{}}0.9859 \\ ±\\ 0.0003\end{tabular}  & \begin{tabular}[c]{@{}l@{}}14.2333 \\ ±\\ 2.467\end{tabular}   \\ \hline
\multirow{4}{*}{\begin{tabular}[c]{@{}l@{}}Retinal\\ Blood\\ Vessel\end{tabular}} & Drive                                                      & \begin{tabular}[c]{@{}l@{}}0.7967 \\ ±\\ 0.0094\end{tabular} & \begin{tabular}[c]{@{}l@{}}0.7018 \\ ±\\ 0.0033\end{tabular} & \begin{tabular}[c]{@{}l@{}}0.9657 \\ ±\\ 0.0011\end{tabular}  & \begin{tabular}[c]{@{}l@{}}0.9311 \\ ±\\ 0.0060\end{tabular}  & \begin{tabular}[c]{@{}l@{}}0.7783 \\ ±\\ 0.0087\end{tabular} & \begin{tabular}[c]{@{}l@{}}0.8339 \\ ±\\ 0.0061\end{tabular}  & \begin{tabular}[c]{@{}l@{}}0.8030 \\ ±\\ 0.0029\end{tabular}  & \begin{tabular}[c]{@{}l@{}}5.3380 \\ ±\\ 0.3828\end{tabular}   \\ \cline{2-10} 
                                                                                  & STARE                                                      & \begin{tabular}[c]{@{}l@{}}0.8001 \\ ±\\ 0.0057\end{tabular} & \begin{tabular}[c]{@{}l@{}}0.7100 \\ ±\\ 0.0046\end{tabular} & \begin{tabular}[c]{@{}l@{}}0.9780 \\ ±\\ 0.0007\end{tabular}  & \begin{tabular}[c]{@{}l@{}}0.9439 \\ ±\\ 0.0034\end{tabular}  & \begin{tabular}[c]{@{}l@{}}0.7931 \\ ±\\ 0.0154\end{tabular} & \begin{tabular}[c]{@{}l@{}}0.8296 \\ ±\\ 0.0125\end{tabular}  & \begin{tabular}[c]{@{}l@{}}0.8083 \\ ±\\ 0.0035\end{tabular}  & \begin{tabular}[c]{@{}l@{}}4.2500 \\ ±\\ 0.5622\end{tabular}   \\ \cline{2-10} 
                                                                                  
                                                                                  & \begin{tabular}[c]{@{}l@{}}CHASE\\ \_DB1\end{tabular}      & \begin{tabular}[c]{@{}l@{}}0.8209 \\ ±\\ 0.0041\end{tabular} & \begin{tabular}[c]{@{}l@{}}0.7352 \\ ±\\ 0.0031\end{tabular} & \begin{tabular}[c]{@{}l@{}}0.9820 \\ ±\\ 0.0010\end{tabular}  & \begin{tabular}[c]{@{}l@{}}0.9528 \\ ±\\ 0.0011\end{tabular}  & \begin{tabular}[c]{@{}l@{}}0.8090 \\ ±\\ 0.0029\end{tabular} & \begin{tabular}[c]{@{}l@{}}0.8509 \\ ±\\ 0.0052\end{tabular}  & \begin{tabular}[c]{@{}l@{}}0.8289 \\ ±\\ 0.0024\end{tabular}  & \begin{tabular}[c]{@{}l@{}}4.4133 \\ ±\\ 0.8749\end{tabular}   \\ \cline{2-10} 
                                                                                  & Combined                                                   & \begin{tabular}[c]{@{}l@{}}0.8106 \\ ±\\ 0.0024\end{tabular} & \begin{tabular}[c]{@{}l@{}}0.7178 \\ ±\\ 0.0013\end{tabular} & \begin{tabular}[c]{@{}l@{}}0.9717 \\ ±\\ 0.0009\end{tabular}  & \begin{tabular}[c]{@{}l@{}}0.9400 \\ ±\\ 0.0006\end{tabular}  & \begin{tabular}[c]{@{}l@{}}0.7965 \\ ±\\ 0.0030\end{tabular} & \begin{tabular}[c]{@{}l@{}}0.8393 \\ ±\\ 0.0012\end{tabular}  & \begin{tabular}[c]{@{}l@{}}0.8159 \\ ±\\ 0.0011\end{tabular}  & \begin{tabular}[c]{@{}l@{}}7.3600 \\ ±\\ 1.9200\end{tabular}   \\ \hline
\multirow{3}{*}{\begin{tabular}[c]{@{}l@{}}Colon\\ Polyp\end{tabular}}            & \begin{tabular}[c]{@{}l@{}}Kvasir\\ -SEG\end{tabular}      & \begin{tabular}[c]{@{}l@{}}0.9327 \\ ±\\ 0.0020\end{tabular} & \begin{tabular}[c]{@{}l@{}}0.8923 \\ ±\\ 0.0029\end{tabular} & \begin{tabular}[c]{@{}l@{}}0.9794 \\ ±\\ 0.0033\end{tabular}  & \begin{tabular}[c]{@{}l@{}}0.9694 \\ ±\\ 0.0015\end{tabular}  & \begin{tabular}[c]{@{}l@{}}0.9388 \\ ±\\ 0.0041\end{tabular} & \begin{tabular}[c]{@{}l@{}}0.9435 \\ ±\\ 0.0036\end{tabular}  & \begin{tabular}[c]{@{}l@{}}0.9331 \\ ±\\ 0.0022\end{tabular}  & \begin{tabular}[c]{@{}l@{}}22.6900 \\ ±\\ 3.4528\end{tabular}   \\ \cline{2-10} 
                                                                                  & \begin{tabular}[c]{@{}l@{}}CVC-\\ Clinic\\ DB\end{tabular} & \begin{tabular}[c]{@{}l@{}}0.9605 \\ ±\\ 0.0035\end{tabular} & \begin{tabular}[c]{@{}l@{}}0.9294 \\ ±\\ 0.0049\end{tabular} & \begin{tabular}[c]{@{}l@{}}0.9934 \\ ±\\ 0.0007\end{tabular}  & \begin{tabular}[c]{@{}l@{}}0.9889 \\ ±\\ 0.0009\end{tabular}  & \begin{tabular}[c]{@{}l@{}}0.9619 \\ ±\\ 0.0056\end{tabular} & \begin{tabular}[c]{@{}l@{}}0.9630 \\ ±\\ 0.0023\end{tabular}  & \begin{tabular}[c]{@{}l@{}}0.9608 \\ ±\\ 0.0034\end{tabular}  & \begin{tabular}[c]{@{}l@{}}24.5100 \\ ±\\ 2.678\end{tabular}   \\ \cline{2-10} 
                                                                                  & Combined                                                   & \begin{tabular}[c]{@{}l@{}}0.9451 \\ ±\\ 0.0012\end{tabular} & \begin{tabular}[c]{@{}l@{}}0.9085 \\ ±\\ 0.0014\end{tabular} & \begin{tabular}[c]{@{}l@{}}0.9829 \\ ±\\ 0.0010\end{tabular}  & \begin{tabular}[c]{@{}l@{}}0.9763 \\ ±\\ 0.0007\end{tabular}  & \begin{tabular}[c]{@{}l@{}}0.9530 \\ ±\\ 0.0016\end{tabular} & \begin{tabular}[c]{@{}l@{}}0.9496 \\ ±\\ 0.0003\end{tabular}  & \begin{tabular}[c]{@{}l@{}}0.9452 \\ ±\\ 0.0011\end{tabular}  & \begin{tabular}[c]{@{}l@{}}49.7700 \\ ±\\ 3.3788\end{tabular}   \\ \hline
\multirow{7}{*}{\begin{tabular}[c]{@{}l@{}}Optic-\\ cup\end{tabular}}             & CRFO                                                       & \begin{tabular}[c]{@{}l@{}}0.8706 \\ ±\\ 0.0145\end{tabular} & \begin{tabular}[c]{@{}l@{}}0.8202 \\ ±\\ 0.0114\end{tabular} & \begin{tabular}[c]{@{}l@{}}0.9966 \\ ±\\ 0.0033\end{tabular}  & \begin{tabular}[c]{@{}l@{}}0.9957 \\ ±\\ 0.0010\end{tabular}  & \begin{tabular}[c]{@{}l@{}}0.8963 \\ ±\\ 0.0204\end{tabular} & \begin{tabular}[c]{@{}l@{}}0.8879 \\ ±\\ 0.0306\end{tabular}  & \begin{tabular}[c]{@{}l@{}}0.8748 \\ ±\\ 0.0103\end{tabular}  & \begin{tabular}[c]{@{}l@{}}12.5333 \\ ±\\ 2.92\end{tabular}   \\ \cline{2-10} 
                                                                                  & DRISHTI                                                    & \begin{tabular}[c]{@{}l@{}}0.9485 \\ ±\\ 0.0045\end{tabular} & \begin{tabular}[c]{@{}l@{}}0.9112 \\ ±\\ 0.0071\end{tabular} & \begin{tabular}[c]{@{}l@{}}0.9990 \\ ±\\ 0.0008\end{tabular}  & \begin{tabular}[c]{@{}l@{}}0.9976 \\ ±\\ 0.0002\end{tabular}  & \begin{tabular}[c]{@{}l@{}}0.9451 \\ ±\\ 0.0063\end{tabular} & \begin{tabular}[c]{@{}l@{}}0.9627 \\ ±\\ 0.0106\end{tabular}  & \begin{tabular}[c]{@{}l@{}}0.9489 \\ ±\\ 0.0048\end{tabular}  & \begin{tabular}[c]{@{}l@{}}24.8533 \\ ±\\ 3.4808\end{tabular}   \\ \cline{2-10} 
                                                                                  & G1020                                                      & \begin{tabular}[c]{@{}l@{}}0.9187 \\ ±\\ 0.0023\end{tabular} & \begin{tabular}[c]{@{}l@{}}0.8642 \\ ±\\ 0.0037\end{tabular} & \begin{tabular}[c]{@{}l@{}}0.9959 \\ ±\\ 0.0001\end{tabular}  & \begin{tabular}[c]{@{}l@{}}0.9975 \\ ±\\ 0.0002\end{tabular}  & \begin{tabular}[c]{@{}l@{}}0.9254 \\ ±\\ 0.0003\end{tabular} & \begin{tabular}[c]{@{}l@{}}0.9280 \\ ±\\ 0.0044\end{tabular}  & \begin{tabular}[c]{@{}l@{}}0.9189 \\ ±\\ 0.0025\end{tabular}  & \begin{tabular}[c]{@{}l@{}}26.2067 \\ ±\\ 1.2349\end{tabular}   \\ \cline{2-10} 
                                                                                  & ORIGA                                                      & \begin{tabular}[c]{@{}l@{}}0.9306 \\ ±\\ 0.0017\end{tabular} & \begin{tabular}[c]{@{}l@{}}0.8826 \\ ±\\ 0.0026\end{tabular} & \begin{tabular}[c]{@{}l@{}}0.9954 \\ ±\\ 0.0024\end{tabular}  & \begin{tabular}[c]{@{}l@{}}0.9969 \\ ±\\ 0.0011\end{tabular}  & \begin{tabular}[c]{@{}l@{}}0.9413 \\ ±\\ 0.0079\end{tabular} & \begin{tabular}[c]{@{}l@{}}0.9352 \\ ±\\ 0.0043\end{tabular}  & \begin{tabular}[c]{@{}l@{}}0.9308 \\ ±\\ 0.0016\end{tabular}  & \begin{tabular}[c]{@{}l@{}}25.0400 \\ ±\\ 2.4802\end{tabular}   \\ \cline{2-10} 
                                                                                  & PAPILA                                                     & \begin{tabular}[c]{@{}l@{}}0.8823 \\ ±\\ 0.0035\end{tabular} & \begin{tabular}[c]{@{}l@{}}0.8236 \\ ±\\ 0.0035\end{tabular} & \begin{tabular}[c]{@{}l@{}} 0.9947 \\ ±\\ 0.0003\end{tabular}  & \begin{tabular}[c]{@{}l@{}}0.9965 \\ ±\\ 0.0001\end{tabular}  & \begin{tabular}[c]{@{}l@{}}0.9020 \\ ±\\ 0.0039\end{tabular} & \begin{tabular}[c]{@{}l@{}}0.8948 \\ ±\\ 0.0059\end{tabular}  & \begin{tabular}[c]{@{}l@{}}0.8830 \\ ±\\ 0.0038\end{tabular}  & \begin{tabular}[c]{@{}l@{}}29.7567 \\ ±\\ 2.5762\end{tabular}   \\ \cline{2-10} 
                                                                                  & REFUGE1                                                    & \begin{tabular}[c]{@{}l@{}}0.9430 \\ ±\\ 0.0027\end{tabular} & \begin{tabular}[c]{@{}l@{}}0.8997 \\ ±\\ 0.0039\end{tabular} & \begin{tabular}[c]{@{}l@{}}0.9969 \\ ±\\ 0.0018\end{tabular}  & \begin{tabular}[c]{@{}l@{}}0.9979 \\ ±\\ 0.0008\end{tabular}  & \begin{tabular}[c]{@{}l@{}}0.9464 \\ ±\\ 0.0051\end{tabular} & \begin{tabular}[c]{@{}l@{}}0.9482 \\ ±\\ 0.0172\end{tabular}  & \begin{tabular}[c]{@{}l@{}}0.9431 \\ ±\\ 0.0026\end{tabular}  & \begin{tabular}[c]{@{}l@{}}31.5867 \\ ±\\ 3.3564\end{tabular}   \\ \cline{2-10} 
                                                                                  & Combined                                                   & \begin{tabular}[c]{@{}l@{}}0.9252 \\ ±\\ 0.0020\end{tabular} & \begin{tabular}[c]{@{}l@{}}0.8763 \\ ±\\ 0.0029\end{tabular} & \begin{tabular}[c]{@{}l@{}}0.9976 \\ ±\\ 0.0011\end{tabular}  & \begin{tabular}[c]{@{}l@{}}0.9974 \\ ±\\ 0.0009\end{tabular}  & \begin{tabular}[c]{@{}l@{}}0.9265 \\ ±\\ 0.0038\end{tabular} & \begin{tabular}[c]{@{}l@{}}0.9402 \\ ±\\ 0.0020\end{tabular}  & \begin{tabular}[c]{@{}l@{}}0.9253 \\ ±\\ 0.0021\end{tabular}  & \begin{tabular}[c]{@{}l@{}}70.3667 \\ ±\\ 14.6573\end{tabular}   \\ \hline
\multirow{7}{*}{\begin{tabular}[c]{@{}l@{}}Optic-\\ Disc\end{tabular}}            & CRFO                                                       & \begin{tabular}[c]{@{}l@{}}0.9772 \\ ±\\ 0.0036\end{tabular} & \begin{tabular}[c]{@{}l@{}}0.9597 \\ ±\\ 0.0062\end{tabular} & \begin{tabular}[c]{@{}l@{}}0.9998 \\ ±\\ 0.0008\end{tabular}  & \begin{tabular}[c]{@{}l@{}}0.9986 \\ ±\\ 0.0002\end{tabular}  & \begin{tabular}[c]{@{}l@{}}0.9818 \\ ±\\ 0.0049\end{tabular} & \begin{tabular}[c]{@{}l@{}}0.9768 \\ ±\\ 0.0101\end{tabular}  & \begin{tabular}[c]{@{}l@{}}0.9789 \\ ±\\ 0.0135\end{tabular}  & \begin{tabular}[c]{@{}l@{}}9.7033 \\ ±\\ 0.9666\end{tabular}   \\ \cline{2-10} 
                                                                                  & DRISHTI                                                    & \begin{tabular}[c]{@{}l@{}}0.9842 \\ ±\\ 0.0009\end{tabular} & \begin{tabular}[c]{@{}l@{}}0.9707 \\ ±\\ 0.0013\end{tabular} & \begin{tabular}[c]{@{}l@{}}0.9998 \\ ±\\ 0.0004\end{tabular}  & \begin{tabular}[c]{@{}l@{}}0.9984 \\ ±\\ 0.0001\end{tabular}  & \begin{tabular}[c]{@{}l@{}}0.9838 \\ ±\\ 0.0016\end{tabular} & \begin{tabular}[c]{@{}l@{}}0.9863 \\ ±\\ 0.0154\end{tabular}  & \begin{tabular}[c]{@{}l@{}}0.9849 \\ ±\\ 0.0007\end{tabular}  & \begin{tabular}[c]{@{}l@{}}20.7533 \\ ±\\ 2.654\end{tabular}   \\ \cline{2-10} 
                                                                                  & G1020                                                      & \begin{tabular}[c]{@{}l@{}} 0.9667 \\ ±\\ 0.0032\end{tabular} & \begin{tabular}[c]{@{}l@{}}0.9394  \\ ±\\ 0.0049\end{tabular} & \begin{tabular}[c]{@{}l@{}}0.9987 \\ ±\\ 0.0001\end{tabular}  & \begin{tabular}[c]{@{}l@{}}0.9979 \\ ±\\ 0.0002\end{tabular}  & \begin{tabular}[c]{@{}l@{}}0.9663 \\ ±\\ 0.0049\end{tabular} & \begin{tabular}[c]{@{}l@{}}0.9700 \\ ±\\ 0.0014\end{tabular}  & \begin{tabular}[c]{@{}l@{}}0.9667 \\ ±\\ 0.0032\end{tabular}  & \begin{tabular}[c]{@{}l@{}}27.3033 \\ ±\\ 7.3422\end{tabular}   \\ \cline{2-10} 
                                                                                  & ORIGA                                                      & \begin{tabular}[c]{@{}l@{}}0.9805 \\ ±\\ 0.0020\end{tabular} & \begin{tabular}[c]{@{}l@{}}0.9628 \\ ±\\ 0.0040\end{tabular} & \begin{tabular}[c]{@{}l@{}}0.9955 \\ ±\\ 0.0057\end{tabular}  & \begin{tabular}[c]{@{}l@{}}0.9963 \\ ±\\ 0.0037\end{tabular}  & \begin{tabular}[c]{@{}l@{}}0.9792 \\ ±\\ 0.0055\end{tabular} & \begin{tabular}[c]{@{}l@{}}0.9828 \\ ±\\ 0.0015\end{tabular}  & \begin{tabular}[c]{@{}l@{}}0.9806 \\ ±\\ 0.0020\end{tabular}  & \begin{tabular}[c]{@{}l@{}}24.7867 \\ ±\\ 4.2306\end{tabular}   \\ \cline{2-10} 
                                                                                  & PAPILA                                                     & \begin{tabular}[c]{@{}l@{}}0.9762 \\ ±\\ 0.0006\end{tabular} & \begin{tabular}[c]{@{}l@{}}0.9558 \\ ±\\ 0.0008\end{tabular} & \begin{tabular}[c]{@{}l@{}}0.9985 \\ ±\\ 0.0001\end{tabular}  & \begin{tabular}[c]{@{}l@{}}0.9973 \\ ±\\ 0.0001\end{tabular}  & \begin{tabular}[c]{@{}l@{}}0.9723 \\ ±\\ 0.0024\end{tabular} & \begin{tabular}[c]{@{}l@{}}0.9825 \\ ±\\ 0.0014\end{tabular}  & \begin{tabular}[c]{@{}l@{}}0.9764 \\ ±\\ 0.0007\end{tabular}  & \begin{tabular}[c]{@{}l@{}}32.6200 \\ ±\\ 7.6245\end{tabular}   \\ \cline{2-10} 
                                                                                  & REFUGE1                                                    & \begin{tabular}[c]{@{}l@{}}0.9791 \\ ±\\ 0.0004\end{tabular} & \begin{tabular}[c]{@{}l@{}}0.9608 \\ ±\\ 0.0004\end{tabular} & \begin{tabular}[c]{@{}l@{}}0.9993 \\ ±\\ 0.0000\end{tabular}  & \begin{tabular}[c]{@{}l@{}}0.9986 \\ ±\\ 0.0001\end{tabular}  & \begin{tabular}[c]{@{}l@{}}0.9787 \\ ±\\ 0.0017\end{tabular} & \begin{tabular}[c]{@{}l@{}}0.9814 \\ ±\\ 0.0013\end{tabular}  & \begin{tabular}[c]{@{}l@{}}0.9792 \\ ±\\ 0.0004\end{tabular}  & \begin{tabular}[c]{@{}l@{}}35.4500 \\ ±\\ 3.0106\end{tabular}   \\ \cline{2-10} 
                                                                                  & Combined                                                   & \begin{tabular}[c]{@{}l@{}}0.9775 \\ ±\\ 0.0005\end{tabular} & \begin{tabular}[c]{@{}l@{}}0.9579 \\ ±\\ 0.0008\end{tabular} & \begin{tabular}[c]{@{}l@{}}0.9990 \\ ±\\ 0.0008\end{tabular}  & \begin{tabular}[c]{@{}l@{}}0.9984 \\ ±\\ 0.0001\end{tabular}  & \begin{tabular}[c]{@{}l@{}}0.9768 \\ ±\\ 0.0008\end{tabular} & \begin{tabular}[c]{@{}l@{}}0.9800 \\ ±\\ 0.0001\end{tabular}  & \begin{tabular}[c]{@{}l@{}}0.9775 \\ ±\\ 0.0005\end{tabular}  & \begin{tabular}[c]{@{}l@{}}146.1033 \\ ±\\ 11.9708\end{tabular}   \\ \hline
\end{longtable}

From the table, it is evident that DeepFusionLab performed well too across segmentation datasets. However, due to a very small number of images in retinal blood vessel datasets, the score was not as high as it was on other datasets. Training time varied mainly due to the number of images and the execution of early stopping. When the image quality is good, the training converges fast, and early stopping executes. But for not-so-good image quality, training does not converge that fast, and so it requires higher training time.

\subsection{Ablation Study of the DeepFusionLab:}
An ablation study is conducted in this subsection to evaluate the contribution of individual components.

Table \ref{tab:Ablation} demonstrates the ablation study by adding or removing various components from the model. Weights are generated from combined datasets, and the classification pathway does not include any blocks except ASPP. Thus, we performed an ablation study on the combined segmentation datasets.
\begin{table}[h!]
\caption{Ablation Study}
\label{tab:Ablation}
\begin{tabular}{|l|llllllllll|}
\hline
Model                                                                                                & \multicolumn{10}{l|}{Combined Dataset and Number of Training Dataset Images}                                                                                                                                                                                                                                                                                                          \\ \hline
\multirow{4}{*}{\begin{tabular}[c]{@{}l@{}}Model \&\\ No of \\ Parameters\end{tabular}}                                                                              & \multicolumn{2}{l|}{Lung}                               & \multicolumn{2}{l|}{\begin{tabular}[c]{@{}l@{}}Retinal\\ Blood\\ Vessel\end{tabular}} & \multicolumn{2}{l|}{\begin{tabular}[c]{@{}l@{}}Colon\\ Polyp\end{tabular}} & \multicolumn{2}{l|}{\begin{tabular}[c]{@{}l@{}}Optic-\\ cup\end{tabular}} & \multicolumn{2}{l|}{\begin{tabular}[c]{@{}l@{}}Optic-\\ Disc\end{tabular}} \\ \cline{2-11} 
                                                                                                     & \multicolumn{2}{l|}{307}                                & \multicolumn{2}{l|}{58}                                                               & \multicolumn{2}{l|}{1289}                                                  & \multicolumn{2}{l|}{2297}                                                 & \multicolumn{2}{l|}{2481}                                                  \\ \cline{2-11} 
                                                                                                     & \multicolumn{1}{l|}{Dice}  & \multicolumn{1}{l|}{IoU}   & \multicolumn{1}{l|}{Dice}                 & \multicolumn{1}{l|}{IoU}                  & \multicolumn{1}{l|}{Dice}            & \multicolumn{1}{l|}{IoU}            & \multicolumn{1}{l|}{Dice}           & \multicolumn{1}{l|}{IoU}            & \multicolumn{1}{l|}{Dice}                      & IoU                       \\ \hline
\begin{tabular}[c]{@{}l@{}}Efficient\\ Net B0+ \\ Decoder \\ 4.1M\end{tabular}                       & \multicolumn{1}{l|}{0.981} & \multicolumn{1}{l|}{0.963} & \multicolumn{1}{l|}{0.792}                & \multicolumn{1}{l|}{0.696}                & \multicolumn{1}{l|}{0.906}           & \multicolumn{1}{l|}{0.865}          & \multicolumn{1}{l|}{0.871}          & \multicolumn{1}{l|}{0.803}          & \multicolumn{1}{l|}{0.925}                     & 0.908                     \\ \hline
\begin{tabular}[c]{@{}l@{}}Efficient\\ Net B0+\\ ASPP+ \\ Decoder\\ 6.2M\end{tabular}                & \multicolumn{1}{l|}{0.982} & \multicolumn{1}{l|}{0.969} & \multicolumn{1}{l|}{0.799}                & \multicolumn{1}{l|}{0.699}                & \multicolumn{1}{l|}{0.910}           & \multicolumn{1}{l|}{0.869}          & \multicolumn{1}{l|}{0.874}          & \multicolumn{1}{l|}{0.806}          & \multicolumn{1}{l|}{0.929}                     & 0.920                     \\ \hline
\begin{tabular}[c]{@{}l@{}}Efficient\\ Net B0+\\ ASPP+\\ Decoder+\\ MFF \\ 6.8M\end{tabular}         & \multicolumn{1}{l|}{0.984} & \multicolumn{1}{l|}{0.971} & \multicolumn{1}{l|}{0.806}                & \multicolumn{1}{l|}{0.703}                & \multicolumn{1}{l|}{0.929}           & \multicolumn{1}{l|}{0.882}          & \multicolumn{1}{l|}{0.882}          & \multicolumn{1}{l|}{0.830}          & \multicolumn{1}{l|}{0.933}                     & 0.921                     \\ \hline
\begin{tabular}[c]{@{}l@{}}Proposed\\ Model +\\ CBAM\\ 8.1M\end{tabular}                             & \multicolumn{1}{l|}{0.986} & \multicolumn{1}{l|}{0.972} & \multicolumn{1}{l|}{0.807}                & \multicolumn{1}{l|}{0.707}                & \multicolumn{1}{l|}{0.945}           & \multicolumn{1}{l|}{0.909}          & \multicolumn{1}{l|}{0.927}          & \multicolumn{1}{l|}{0.877}          & \multicolumn{1}{l|}{0.979}                     & 0.959                     \\ \hline
\begin{tabular}[c]{@{}l@{}}Proposed \\ Model \\ but with \\ Efficient\\ Net B1 \\ 10.2M\end{tabular} & \multicolumn{1}{l|}{0.987} & \multicolumn{1}{l|}{0.974} & \multicolumn{1}{l|}{0.821}                & \multicolumn{1}{l|}{0.718}                & \multicolumn{1}{l|}{0.950}           & \multicolumn{1}{l|}{0.909}          & \multicolumn{1}{l|}{0.931}          & \multicolumn{1}{l|}{0.879}          & \multicolumn{1}{l|}{0.979}                     & 0.961                     \\ \hline
\begin{tabular}[c]{@{}l@{}}Proposed \\ Model\\ 8.12M\end{tabular}                                       & \multicolumn{1}{l|}{0.986} & \multicolumn{1}{l|}{0.973} & \multicolumn{1}{l|}{0.811}                & \multicolumn{1}{l|}{0.718}                & \multicolumn{1}{l|}{0.945}           & \multicolumn{1}{l|}{0.909}          & \multicolumn{1}{l|}{0.925}          & \multicolumn{1}{l|}{0.877}          & \multicolumn{1}{l|}{0.978}                     & 0.958                     \\ \hline
\end{tabular}
\end{table}

From the ablation study, we see that for the lung and retinal blood vessels, all the models performed well. As the retinal blood vessel has only 58 training images, adding the CBAM attention module introduced overfitting and reduced performance.  
When the number of training images is higher, the proposed model has a performance gain over its variants with fewer components. But even in this case, adding the CBAM module does not enhance the performance to a great extent.
Finally, we checked with EfficientNet-B1 as the encoder. This increased the number of parameters from 8.12M to 10.2M, but did not increase the performance that much.

So considering all these, the proposed architecture was chosen that has a decent number of parameters but does not compromise performance, and works well on both small-sized and large-sized datasets.

\subsection{Overall Framework Performance}

The proposed approach uniquely combines natural language understanding, dynamic task planning, and a modular CNN (DeepFusionLab) for flexible, user-driven medical image analysis without requiring full retraining for new tasks. Unlike most prior studies that focus on isolated components or fixed-task solutions, MedPrompt addresses the distinct research question of building a scalable, adaptable system for prompt-driven multi-task inference. Therefore, we evaluate MedPrompt's system-level performance through metrics such as end-to-end correctness and inference latency, which capture its unique capabilities. To assess its performance, we define a set of custom metrics that evaluate the correctness of each component within the pipeline:

\begin{itemize}
     \item \textbf{Intention Correctness:} This measures whether the LLM can accurately determine the intent of each task, specifically, whether it is a classification or segmentation task.
     
     \item \textbf{Target Correctness:} This evaluates the LLM's ability to correctly identify and normalize the target organ or disease mentioned in the user query. For instance, if the user refers to the "pulmonary parenchyma region," the framework should infer that the target corresponds to the "lung region."

     \item \textbf{Weight Correctness:} This assesses whether the LLM can select the most appropriate model weight for each task. If no suitable weight exists, the framework should recognize this rather than assigning one arbitrarily.

    \item \textbf{Condition and Dependency Correctness:} This captures the LLM’s ability to detect inter-task dependencies and execution conditions. For example, if a user requests lung segmentation only in the presence of pneumonia, the framework must understand that "lung segmentation" is dependent on the "existence of pneumonia" and execute it conditionally.

    \item \textbf{Overall Correctness:} This metric reflects how accurately the final output of the pipeline addresses the user's query. For example, if a user asks, “Does the image show signs of tuberculosis?”, this metric evaluates whether the system returns a correct and coherent response.

The five metrics above are mainly used to check if the JSON structure is correct. But they also help us see if the final output is correct. In our tests, when the JSON had the right intent, target, and other details, the final output also got them right. So instead of showing these metrics in a different table, we included them with the main results.

    \item \textbf{Average Time:} Given the pipeline's complexity—including tasks like weight selection and dependency resolution—this metric measures the average time required to respond to a user query. It helps determine whether the framework is viable for real-time or interactive applications. We conducted the framework evaluation through Kaggle's CPU. So the time given in the Table refers to the average time required for final response generation while using a CPU. 

\end{itemize}

Table \ref{tab:MedP} demonstrates the evaluation of the proposed framework. 

\begin{table}[h!]
\caption{Performance of MedPrompt}
\label{tab:MedP}
\begin{tabular}{|l|l|lllll|l|l|}
\hline
\multirow{2}{*}{\begin{tabular}[c]{@{}l@{}}Question\\ Type\end{tabular}} & \multirow{2}{*}{\begin{tabular}[c]{@{}l@{}}Number \\ of \\ Questions\end{tabular}} & \multicolumn{5}{l|}{Correctly Identified}                                                                                                                                                                                        & \multirow{2}{*}{\begin{tabular}[c]{@{}l@{}}Overall\\ Correct\\ ness\end{tabular}} & \multirow{2}{*}{\begin{tabular}[c]{@{}l@{}}Average\\ Time\\ (CPU)\end{tabular}} \\ \cline{3-7}
                                                                         &                                                                                    & \multicolumn{1}{l|}{Intention} & \multicolumn{1}{l|}{Target}  & \multicolumn{1}{l|}{Weight}  & \multicolumn{1}{l|}{\begin{tabular}[c]{@{}l@{}}Condi\\ tion\end{tabular}} & \begin{tabular}[c]{@{}l@{}}Depen\\ dency\end{tabular} &                                                                                   &                                                                                 \\ \hline
\multirow{2}{*}{Simple}                                                  & \multirow{2}{*}{50}                                                                & \multicolumn{1}{l|}{50/50}     & \multicolumn{1}{l|}{49/50}   & \multicolumn{1}{l|}{49/50}   & \multicolumn{1}{l|}{\multirow{2}{*}{Null}}                                & \multirow{2}{*}{Null}                                 & 49/50                                                                             & \multirow{2}{*}{2.11s}                                                          \\ \cline{3-5} \cline{8-8}
                                                                         &                                                                                    & \multicolumn{1}{l|}{100\%}     & \multicolumn{1}{l|}{98\%}    & \multicolumn{1}{l|}{98\%}    & \multicolumn{1}{l|}{}                                                     &                                                       & 98\%                                                                              &                                                                                 \\ \hline
\multirow{2}{*}{Complex}                                                 & \multirow{2}{*}{50}                                                                & \multicolumn{1}{l|}{114/115}   & \multicolumn{1}{l|}{114/115} & \multicolumn{1}{l|}{112/115} & \multicolumn{1}{l|}{57/60}                                                & 57/60                                                 & 111/115                                                                           & \multirow{2}{*}{2.77s}                                                          \\ \cline{3-8}
                                                                         &                                                                                    & \multicolumn{1}{l|}{99.13\%}   & \multicolumn{1}{l|}{99.13\%} & \multicolumn{1}{l|}{97.39\%} & \multicolumn{1}{l|}{95\%}                                              & 95\%                                               & 96.52\%                                                                            &                                                                                 \\ \hline
\multirow{2}{*}{Overall}                                                 & \multirow{2}{*}{100}                                                               & \multicolumn{1}{l|}{164/165}   & \multicolumn{1}{l|}{163/165} & \multicolumn{1}{l|}{161/165} & \multicolumn{1}{l|}{57/60}                                                & 57/60                                                 & 160/165                                                                           & \multirow{2}{*}{2.44s}                                                           \\ \cline{3-8}
                                                                         &                                                                                    & \multicolumn{1}{l|}{99.39\%}   & \multicolumn{1}{l|}{98.79\%} & \multicolumn{1}{l|}{97.58\%} & \multicolumn{1}{l|}{95\%}                                              & 95\%                                               & 96.97\%                                                                           &                                                                                 \\ \hline
\end{tabular}
\end{table}

For evaluation, we considered both simple and complex prompts. Simple prompts indicate queries that contain only one task with no dependency, like:

-Delimit the pulmonary parenchyma region from the provided thoracic radiograph

-Evaluate this chest radiograph for indications of *pulmonary Mycobacterium tuberculosis* infection.

-Examine this hematologic image for indications of *acute lymphoblastic leukemia

Complex questions indicate queries with multiple tasks, conditions, and dependencies. Some questions that we used for evaluation are:

-Assess for viral etiology of pneumonia. Upon confirmation, segment pulmonary structures for severity quantification. (Simple language: Check for viral pneumonia. If confirmed, segment the lungs for further assessment.)

-Determine if SARS-CoV-2 pneumonia is present. If confirmed, assess for concurrent Mycobacterial pulmonary involvement. (Simple language: Classify this radiograph for COVID. If confirmed, check for any signs of co-existing tuberculosis.)

-From this retinal fundus image, assess for optic neuropathy due to elevated intraocular pressure. If found, delineate the optic cup to support diagnosis. (Simple language: From the fundus image, check for glaucoma. If detected, segment the optic cup for visual confirmation.)

From our evaluation, the framework can very easily understand simple queries with almost perfect accuracy. However, most of the incorrectness (3\%) comes from unclear or complex condition-based questions. One question that the LLM failed to answer is:

-Assess this thoracic radiograph for parenchymal opacification. If inconclusive, screen for viral lower respiratory tract infection.

From the performance evaluation, it is evident that MedPrompt is suitable for real-life applications, with a higher understanding of both the users' simple and complex prompts.

The system is designed to effectively execute even complex plans generated by the LLM. However, when the LLM identifies that a query falls outside its knowledge and database scope, it instructs the system to skip the corresponding task.

\subsection{Impact of LLM Selection}

According to our observations, the choice of LLM primarily affects response time rather than output accuracy. Without a knowledge base, prompt processing averaged 1 second. Using one for enhanced target identification increased latency — Llama 4 took around 2.5 seconds, while Gemma-2, Qwen ranged from 2–5 seconds.

Since the task involves structured, prompt-guided output—not creative generation—all tested LLMs (Llama, Gemma, Qwen) produced similar results. In real-world clinical use, doctors rarely use complex synonyms, reducing the need for heavy knowledge-based normalization. Thus, lighter models or simpler systems may be preferable for time-sensitive settings.

\subsection{Comparison}
To our best understanding, MedPrompt \& its mechanisms are fundamentally different from existing works, making it not justified to compare MedPrompt with other models. Instead, we focus on evaluating the core vision component, DeepFusionLab, against existing segmentation and classification models to ensure a fair and relevant comparison.

\subsubsection{Classification Comparison of DeepFusionLab:}
Table \ref{tab:Cls_Cmp} compares DeepFusionLab with other studies over classification datasets.

\begin{table}[h!]
\caption{Comparison Over Classification Dataset}
\label{tab:Cls_Cmp}
\begin{tabular}{|l|l|l|l|l|}
\hline
Objective                                                                                         & Dataset                                                                                    & Work                                                                          & Accuracy (\%) & f1(\%) \\ \hline
\multirow{5}{*}{Tuberculosis}                                                                     & \multirow{3}{*}{CXR\_Dataset}                                                              & Rahman et al. \citep{rahman2020reliable}                                                                & 98.6          & -      \\ \cline{3-5} 
                                                                                                  &                                                                                            & Huy et al. \citep{huy2023improved}                                                                   & 98.8          & 98.8   \\ \cline{3-5} 
                                                                                                  &                                                                                            & This study                                                                    & \textbf{99.95}         & \textbf{99.23}  \\ \cline{2-5} 
                                                                                                  & \multirow{2}{*}{Shenzhen}                                                                  & Rajaraman et al. \citep{rajaraman2021chest}                                                             & 88.79         & 88.73  \\ \cline{3-5} 
                                                                                                  &                                                                                            & This study                                                                    & \textbf{89.72}         & \textbf{89.71}  \\ \hline
\multirow{3}{*}{Malaria}                                                                                 & \multirow{3}{*}{\begin{tabular}[c]{@{}l@{}}NIH Malaria\\ Dataset\end{tabular}}             & Boit et al. \citep{boit2024efficient}                                                                  & 97.68         & 97.65  \\ \cline{3-5} 
                                                                                                  &                                                                                            & Madhu et al. \citep{madhu2023intelligent}                                                                 & \textbf{99.35}         & \textbf{99.36}  \\ \cline{3-5} 
                                                                                                  &                                                                                            & This study                                                                    & 97.76         & 97.76  \\ \hline
\multirow{3}{*}{Leukemia}                                                                         & \multirow{3}{*}{\begin{tabular}[c]{@{}l@{}}C\_NMC\_\\ Leukemia\end{tabular}}               & Jawahar et al. \citep{jawahar2024attention}                                                               & 91.98         & 96     \\ \cline{3-5} 
                                                                                                  &                                                                                            & Sampathila et al. \citep{sampathila2022customized}                                                             & 95.54         & 95.43  \\ \cline{3-5} 
                                                                                                  &                                                                                            & This study                                                                    & \textbf{98.08}         & \textbf{97.77}  \\ \hline
\multirow{4}{*}{Glaucoma}                                                                         & \multirow{2}{*}{Rim-One}                                                                   & Shoukat et al. \citep{shoukat2023automatic}                                                                 & \textbf{96.15}            & \textbf{97}  \\ \cline{3-5} 
                                                                                                  &                                                                                            & This study                                                                    & 90.31         & 89.60  \\ \cline{2-5} 
                                                                                                  & \multirow{2}{*}{Refuge-2}                                                                  & Neto et al.\citep{neto2022evaluations}                                                                & 97         & -     \\ \cline{3-5} 
                                                                                                  &                                                                                            & This study                                                                    & \textbf{97.50}         & \textbf{92.49}  \\ \hline
\multirow{3}{*}{\begin{tabular}[c]{@{}l@{}}COVID,\\ Viral Pneumonia,\\ Lung Opacity\end{tabular}} & \multirow{3}{*}{\begin{tabular}[c]{@{}l@{}}Covid 19\\ Radiography \\ Dataset\end{tabular}} & \begin{tabular}[c]{@{}l@{}}Yen et al. \citep{yen2024lightweight}\\ (3 Class Classification)\end{tabular} & 97.03         & 97.03  \\ \cline{3-5} 
                                                                                                  &                                                                                            & Chauhan et al. \cite{chauhan2024detection}                                                                & \textbf{98.91}         & \textbf{97.86}  \\ \cline{3-5} 
                                                                                                  &                                                                                            & This study (All Classes)                                                      & 95.80         & 96.45  \\ \hline
\end{tabular}
\end{table}

\subsubsection{Segmentation Comparison of DeepFusionLab:}
The table compares DeepFusionLab with other studies over segmentation datasets.

\begin{table}[h!]
\begin{tabular}{|l|l|l|l|l|}
\hline
Objective                             & Dataset                       & Work                       & IoU (\%)                    & Dice (\%)                  \\ \hline
\multirow{6}{*}{Lung}                 & \multirow{3}{*}{JSRT}         & Liu et al. \citep{liu2022automatic}                 & 95.80                   & 97.90                  \\ \cline{3-5} 
                                      &                               & Yuan et al. \citep{yuan2024leveraging}               & 94.90                   & 97.40                  \\ \cline{3-5} 
                                      &                               & This study                 & \textbf{97.11}                   & \textbf{98.47}                  \\ \cline{2-5} 
                                      & \multirow{3}{*}{Montgomery}   & Liu et al. \citep{liu2022automatic}                 & 95.5                    & 97.4                   \\ \cline{3-5} 
                                      &                               & Yuan et al. \citep{yuan2024leveraging}                & 96.1                    & 98.0                   \\ \cline{3-5} 
                                      &                               & This study                 & \textbf{96.8}                    & \textbf{98.3}                   \\ \hline
\multirow{9}{*}{Retinal Blood Vessel} & \multirow{3}{*}{Drive}        & Liu et al. \citep{liu2024deep}                  & $\sim$                  & 83.3                   \\ \cline{3-5} 
                                      &                               & Rui et al. \citep{yang2024enhancing}                   & 72.34                   & 83.94                  \\ \cline{3-5} 
                                      &                               & This study                 & \textbf{70.18}                   & \textbf{79.67}                  \\ \cline{2-5} 
                                      & \multirow{3}{*}{STARE}        & Liu et al. \citep{liu2024deep}                 & $\sim$                  & 86.6                   \\ \cline{3-5} 
                                      &                               & Rui et al.  \citep{yang2024enhancing}                & 73.53                   & 85.2                   \\ \cline{3-5} 
                                      &                               & This study                 & \textbf{71.0}                    & \textbf{80.0}                   \\ \cline{2-5} 
                                      & \multirow{3}{*}{CHASE\_DB1}   & Liu et al. \citep{liu2024deep}                & $\sim$                  & 81.6                   \\ \cline{3-5} 
                                      &                               & Rui et al. \citep{yang2024enhancing}                 & 70.59                   & 80.4                   \\ \cline{3-5} 
                                      &                               & This study                 & \textbf{73.52}                   & \textbf{82.09}                  \\ \hline
\multirow{6}{*}{Colon-Polyp}          & \multirow{3}{*}{Kvasir-SEG}   & Xue et al. \citep{xue2024lighter}                 & 89.1                    & 93.9                   \\ \cline{3-5} 
                                      &                               & Li et al. \citep{li2025cfformer}                 & 86.3                    & 91.9                   \\ \cline{3-5} 
                                      &                               & This study                 & \textbf{89.23}                   & \textbf{93.27}                  \\ \cline{2-5} 
                                      & \multirow{3}{*}{CVC-ClinicDB} & Xue et al. \citep{xue2024lighter}                  & 90.2                    & 94.7                   \\ \cline{3-5} 
                                      &                               & Li et al. \citep{li2025cfformer}                   & 88.7                    & 93.9                   \\ \cline{3-5} 
                                      &                               & This study                 & \textbf{92.94}                   & \textbf{96.05}                  \\ \hline
\multirow{6}{*}{optic-cup}            & \multirow{2}{*}{DRISHTI}      & Guo et al. \citep{guo2022joint}                &$\sim$                  & 92.7                   \\ \cline{3-5} 
                                      &                               & This study                 & \textbf{82.02}                   & \textbf{87.06} \\ \cline{2-5}
                                      & \multirow{3}{*}{REFUGE1}      & Yi et al. \citep{yi2023c2ftfnet}               &$\sim$ & 90.82 \\ \cline{3-5} 
                                      &                               & Bhattacharya et al. \citep{bhattacharya2023py}       & $\sim$                  & 93.85                  \\ \cline{3-5} 
                                      &                               & This study                 & \textbf{89.97}                   & \textbf{94.30}                  \\ \hline
\multirow{8}{*}{optic-disc}           & \multirow{3}{*}{DRISHTI}      & Guo et al. \citep{guo2022joint}                & 96.57                   & 98.24                  \\ \cline{3-5} 
                                      &                               & Bhattacharya et al. \citep{bhattacharya2023py}       & 94.37                   & 97.10                  \\ \cline{3-5} 
                                      &                               & This study                 & \textbf{97.07}                   & \textbf{98.42}                  \\ \cline{2-5} 
                                      & \multirow{2}{*}{ORIGA}        & Raza et al. \citep{mahmood2025data}               & 88.0                    & 91.0                   \\ \cline{3-5} 
                                      &                               & This study                 & \textbf{96.28}                   & \textbf{98.05}                  \\ \cline{2-5} 
                                      & \multirow{3}{*}{REFUGE1}      & Bhattacharya et al. \citep{bhattacharya2023py}        & 93.23                   & 96.48                  \\ \cline{3-5} 
                                      &                               & Guo et al. \citep{guo2022joint}                & 95.8                    & $\sim$                 \\ \cline{3-5} 
                                      &                               & This study                 & \textbf{96.08}                   & \textbf{97.91}                  \\ \hline
\end{tabular}
\end{table}

\subsubsection{Discussion on Comparison:}
As a model that performs both segmentation and classification on a diverse range of datasets, DeepFusionLab is set for a specific image shape (256,256). Various models take various shapes of input, such as Drive, Stare datasets contain input sizes of (565 × 584) and (700 × 605). So, compared to some other works on these datasets, DeepFusionLab did not perform exceptionally well, but the model still performed very well. Besides adding some more blocks or a better backbone, like EfficientNet B2, might enhance performance in datasets like the COVID-19 Radiography Dataset, but they will increase the number of parameters and might introduce overfitting in datasets like Drive or Stare.

So, considering all this, DeepFusionLab performed well across a diverse range of datasets. 

\subsection{Demonstration}
Figure \ref{JSON} illustrates the generated structured format output by LLM.

\begin{figure}[h]
    \centering
    \includegraphics[width=1\textwidth]{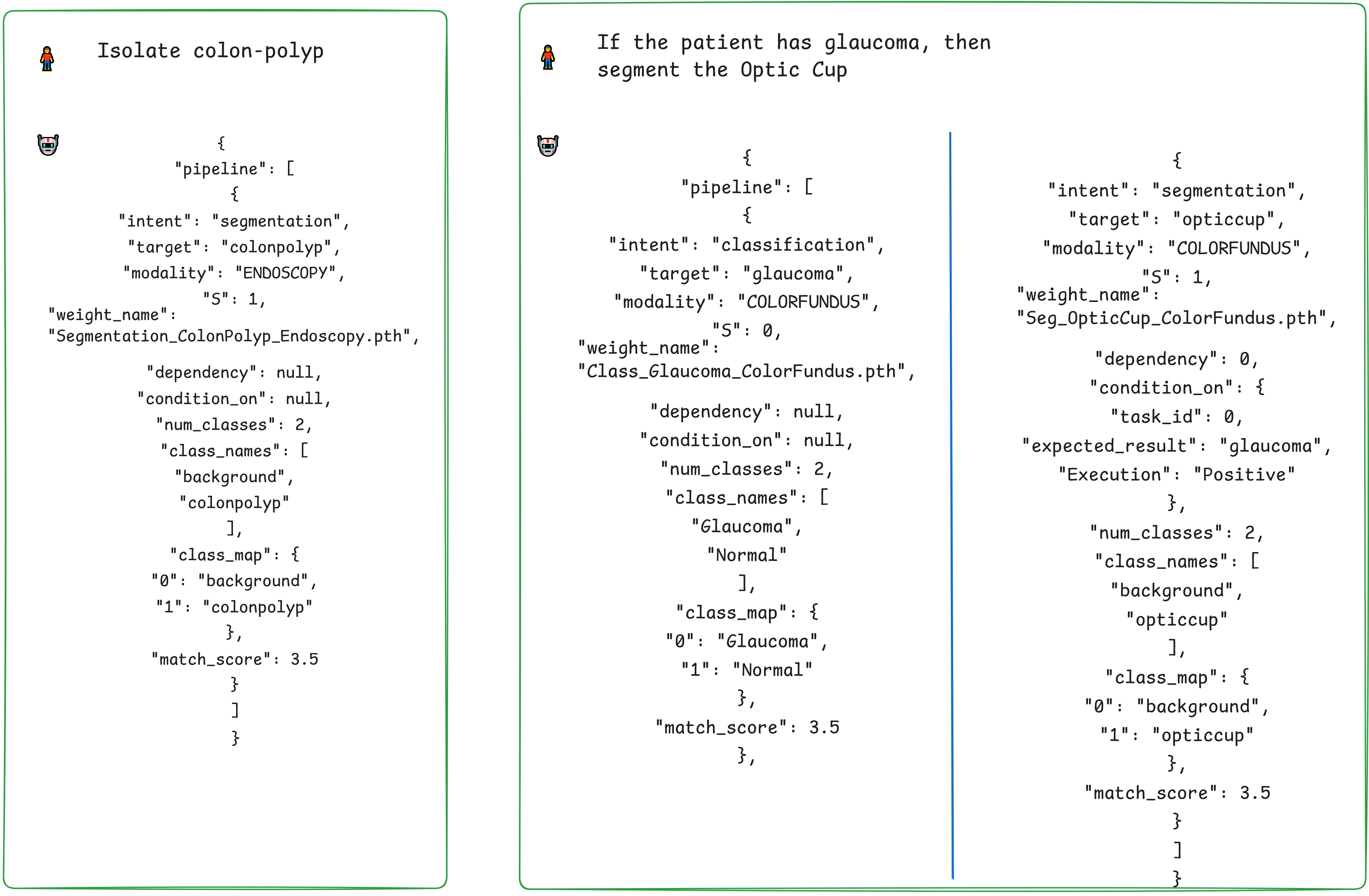}
    \caption{Generated Structured Output by LLM}
    \label{JSON}
\end{figure}

\begin{figure}[h]
    \centering
    \includegraphics[width=1\textwidth]{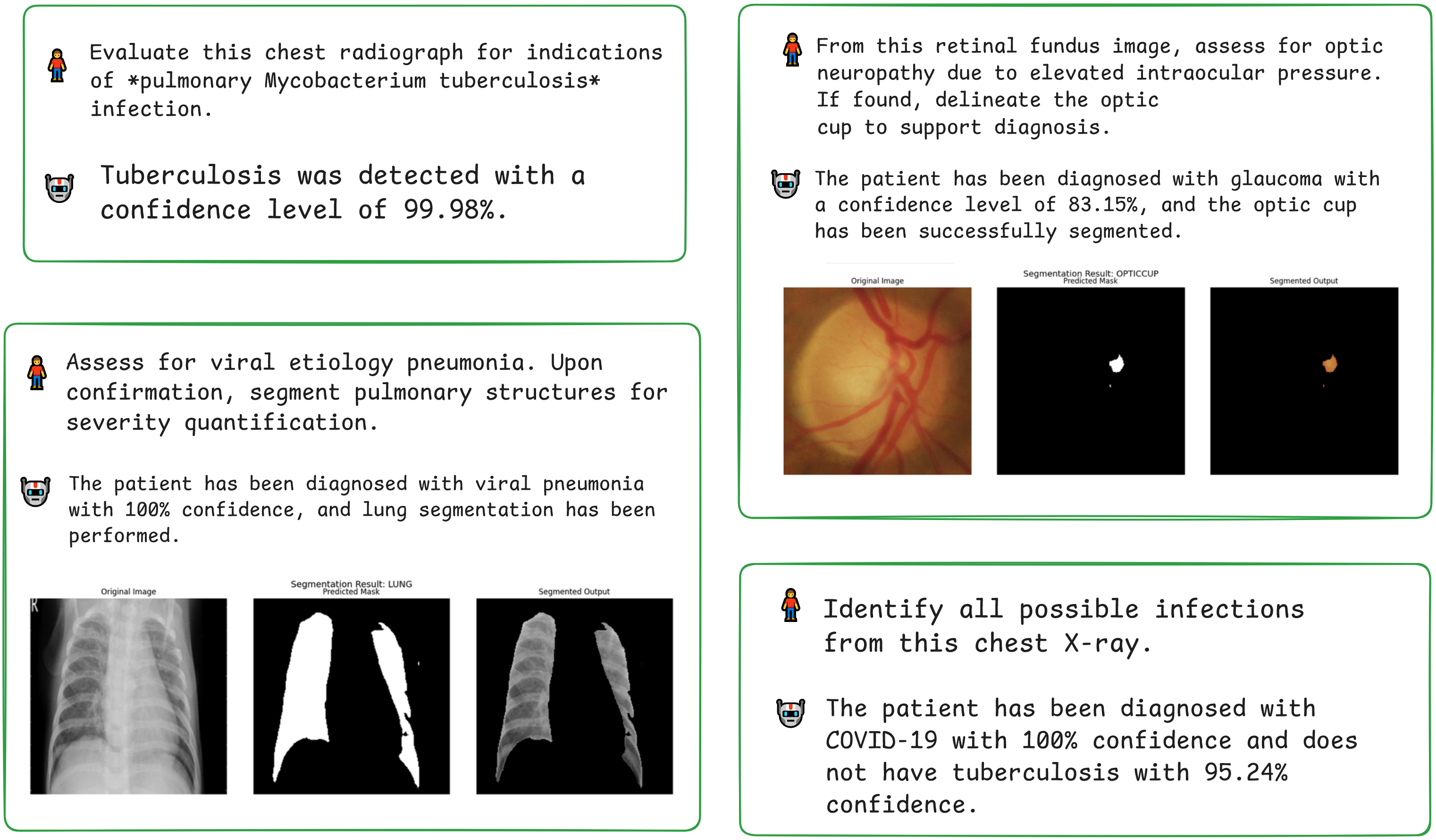}
    \caption{Demonstration of MedPrompt's Question-Answering}
    \label{Medemo}
\end{figure}

Figure \ref{Medemo} demonstrates question answering through MedPrompt.

\section{Conclusion}
\label{con}
In this work, we introduced MedPrompt, which brings together a large language model and a modular CNN to perform medical image classification and segmentation based on natural language instructions. By separating the task planning from the image analysis, MedPrompt allows users to describe what they want using prompts, and the system dynamically selects the right model weights to carry out the task. Our experiments on 19 public datasets show that MedPrompt performs well across 12 tasks and 5 imaging modalities, with high accuracy and low response time, making it suitable for near real-time use. The modular design also makes it easy to add new tasks—only the relevant weights need to be trained and added, without changing the full system. However, the current use is limited to a small set of tasks and imaging types. In the future, we plan to expand the trained weights and include 3D images like CT or MRI. Overall, MedPrompt is a step toward more adaptable and user-friendly medical AI tools that can respond to natural language and work across a wide range of clinical tasks.

\section{Data Availability}
All the processed datasets, training scripts, notebook with evaluation scores, generated model weights, and the main notebook will be made publicly available at: \href{https://github.com/ShadmanSobhan/MedPrompt}{MedPrompt GitHub Repository} upon the publication of this work.

A live demonstration is available on \href{https://youtu.be/f4sTebbGW8M}{Youtube}.

\bibliographystyle{elsarticle-num-names} 

\bibliography{Main}

\end{document}